\documentclass[10pt,journal,compsoc]{IEEEtran}

\usepackage[numbers,sort&compress]{natbib}
\usepackage{microtype}
\usepackage{graphicx}
\usepackage{booktabs}
\usepackage{multirow}
\usepackage{placeins}
\usepackage{amsmath}
\usepackage{amssymb}
\usepackage{mathtools}
\usepackage{amsthm}
\usepackage{algorithm}
\usepackage{algorithmic}
\usepackage[hidelinks]{hyperref}
\usepackage[capitalize,noabbrev]{cleveref}

\theoremstyle{plain}
\newtheorem{proposition}{Proposition}
\theoremstyle{definition}
\newtheorem{definition}{Definition}

\newcommand{\ours}{GraphFlowFM}
\newcommand{\namecite}[2][]{\ifx\relax#1\relax~\cite{#2}\else~\cite[#1]{#2}\fi}

\begin{document}

\title{Versatile Graph Foundation Models from Frozen Time-Series Foundation Models via Parallel Flows}

\author{Jielong~Yang and Feng~Ji%
\IEEEcompsocitemizethanks{
\IEEEcompsocthanksitem J. Yang and F. Ji are with Nanyang Technological University, Singapore.
E-mail: jyang022@e.ntu.edu.sg, jifeng@ntu.edu.sg
\IEEEcompsocthanksitem J. Yang and F. Ji contributed equally. Corresponding author: J. Yang.}}

\markboth{Preprint}%
{Yang and Ji: Versatile Graph Foundation Models from Frozen Time-Series Foundation Models}

\IEEEtitleabstractindextext{%
\begin{abstract}
Graph foundation models aim to serve unseen graphs across tasks, feature spaces, and label sets.
Most current models learn how to represent a new graph by pre-training on other graphs.
Because graph data are scarce and differ across domains in features and structures, what they learn
may not transfer to a graph from an unseen domain.
We show that a frozen time-series foundation model, pre-trained on abundant time series, can instead
serve as a graph foundation model without graph pre-training and without any gradient update on the
target graph.
Such a model reads ordered multivariate sequences, but a graph has no time axis,
and covering all its edges with sequences is nontrivial because a node can have many neighbors.
We therefore propose \ours{}, which decomposes a graph into parallel flows, sets of vertex-disjoint
paths that jointly cover every edge, and reads each path as a multivariate time series with the nodes
as time steps.
A frozen TimesFM encodes the paths with causal attention, which gives each node a self token and a
context token, and a closed-form ridge readout weighs these tokens by their alignment with the
labels.
In node classification, \ours{} attains the best average rank on 21 datasets among graph
foundation models published in 2025 and 2026, ranks first on 17 datasets, and leads on six of the seven
heterophilous datasets.
It also exceeds pre-trained graph foundation models in link prediction on all six datasets and ranks
second in molecule classification, behind only a model fine-tuned after pre-training on molecules.
\end{abstract}

}

\maketitle
\IEEEdisplaynontitleabstractindextext
\IEEEpeerreviewmaketitle

\section{Introduction}
\label{sec:intro}

A graph foundation model is a single model that serves unseen graphs across tasks, feature
spaces, and label sets \citep{wang2025gfmsurvey,zhao2025graphany}.
The prevailing assumption is that such a model can be applied to a new graph only after it has been
pre-trained on graph data, real or synthetic
\citep{liu2024ofa,zhao2025graphany,choi2026nodepfn}.

\begin{figure}[t]
  \centering
  \includegraphics[width=\columnwidth]{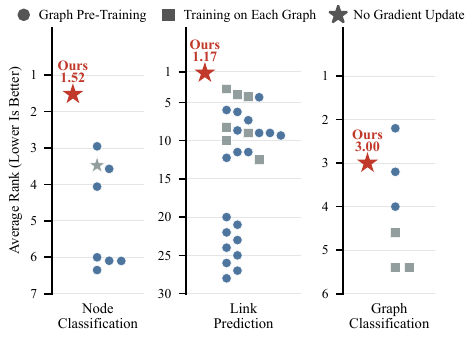}
  \caption{Average rank of every compared method on three tasks, one panel per task with its own rank scale.
  Node classification covers the 21 datasets of \cref{tab:main} and every reported variant, link
  prediction six datasets, and graph classification five molecule benchmarks. Lower is better.
  Our method \ours{} ranks first on node classification and link prediction. On graph classification it ranks second (3.00) behind GIT-S
  (2.20), but GIT-S is pre-trained on molecules and fine-tuned on each task, while the encoder of
  \ours{} has never seen a molecule and makes no gradient update.}
  \label{fig:teaser}
\end{figure}

Three families of graph foundation models follow this assumption, and each has a specific gap.
The first family pre-trains on corpora of many graphs and adapts to a new graph by prompt tuning or
fine-tuning \citep{liu2024ofa,wang2024gft,zhao2024gcope,yu2025samgpt,wang2025mdgfm,sun2025riemanngfm,wang2025git,kong2025gofa}.
These models need text attributes or feature alignment to bridge feature spaces, and they still
update parameters on the target graph.
The second family trains a model once on a single graph and applies it to new graphs without
retraining it \citep{zhao2025graphany,finkelshtein2025equivariance,zhao2026viewspace}.
These models still learn their weights from graph data, and their representations aggregate neighbors.
On the heterophilous Roman-empire graph, where only 5\% of edges join nodes of the same class, the second family reaches 64.1\% to 70.5\% accuracy
\citep{zhao2025graphany,finkelshtein2025equivariance,zhao2026viewspace}.
The third family pre-trains on synthetic graphs drawn from a designed prior
\citep{choi2026nodepfn,eremeev2026graphpfn}, which moves pre-training from real graphs to
generated data, but the graph prior must still be designed and trained.
All three families therefore learn from other graphs how to represent a new graph.
Because graph data are scarce and differ across domains in feature spaces and structures, what is learned from a limited set of pre-training graphs may not
transfer to a graph from an unseen domain.

\emph{These challenges motivate an alternative: reusing knowledge learned from abundant non-graph data,
without requiring graph-specific pre-training.}
A recent method follows this route with a tabular foundation model \citep{hollmann2025tabpfn}
applied to node features aggregated over neighbors \citep{eremeev2025g2tfm,hayler2025tag}.
However, the graph enters only through neighborhood averages and structural encodings computed
before the model, so the neighborhood of a node reaches the model as a fixed summary.
Such averages can obscure differences among neighbors, for example on heterophilous graphs, where
neighbors often belong to different classes.
The approach also addresses node property prediction only.

Time-series foundation models have properties that could mitigate both the dependence on graph
pre-training and the limits of the tabular approach.
They need no graph data, since they are pre-trained on billions of time points from many domains
\citep{das2024timesfm,woo2024moirai,ansari2024chronos,shi2025timemoe,liu2025sundial}.
Some of them also attend jointly across a varying number of variates whose meaning is not fixed in
advance \citep{woo2024moirai,liu2025moiraimoe,jain2026timesfm3}, so graphs with different feature
spaces can be given to one encoder once their features are converted to a common number of
variates.
Unlike a tabular model, they read ordered sequences and give a token to every position, so the
context of a node is encoded by the model rather than reduced to a fixed summary before it.
However, a graph cannot be fed to a time-series foundation model directly, since it has no time
axis.
A path through a graph is an ordered sequence of nodes, but a single path or a set of random walks
may miss edges, because a node can have many neighbors.
The parallel flows that we proposed in \citep{ji2020gfcn} decompose any graph
without training into sets of vertex-disjoint paths that jointly cover every edge.
This observation leads to the question of this paper.
\emph{Can a frozen time-series foundation model serve as a graph foundation model once the graph is
decomposed into parallel flows?}

We answer this question with \ours{}, shown in \cref{fig:method}.
\ours{} decomposes a graph into parallel flows, sets of vertex-disjoint paths taken from
depth-first search forests that together cover every edge.
Each path becomes two multivariate time series.
In the feature series, the 32 variates are the node features compressed to 32 dimensions.
In the structure series, the eight variates are the node degree and random-walk return
probabilities.
A frozen TimesFM-3 \citep{jain2026timesfm3} encodes every series, and causal attention gives each node
a self token and a context token that depends only on the part of the path before the node.
\ours{} averages these tokens over all flows.
A closed-form ridge readout weighs the feature and structure tokens by their alignment with the
training labels and combines kernel candidates by validation accuracy.
No parameter is pre-trained on graphs.
On the target graph, the encoder stays frozen, and only the readout is fitted to the labels, in
closed form without any gradient update.

Our contributions are as follows.
\begin{itemize}
  \item \textbf{Graph foundation models from time-series foundation models.}
  Most graph foundation models are pre-trained on graphs or retrained on each target graph, and
  many are designed for a single task.
  We show that a frozen time-series foundation model can serve as a graph foundation model for node
  classification, link prediction, and graph classification, without graph pre-training and without
  any gradient update.
  \item \textbf{Parallel flows as a sequence view of graphs.}
  A time-series model reads ordered sequences, whereas a graph has no time axis.
  We therefore decompose each graph into parallel flows, vertex-disjoint paths that jointly cover
  every edge, so that every path becomes a multivariate time series and causal attention keeps the
  ordered context of a node instead of averaging its neighbors.
  The decomposition needs no labels and no learnable parameters, so it applies unchanged to any graph.
  \item \textbf{State of the art on three tasks.}
  With one frozen encoder for every dataset and task, \ours{} attains the best average rank
  among 2025 and 2026 graph foundation models on 21 node classification datasets and the highest
  ROC-AUC on five of six link prediction datasets, and it ranks second in molecule classification
  behind a model pre-trained on molecules.
\end{itemize}
\cref{fig:teaser} summarizes these results by the average rank of every compared method on each
task.

\def\methodfigearly{}\begin{figure*}[t]
  \centering
  \includegraphics[width=\textwidth]{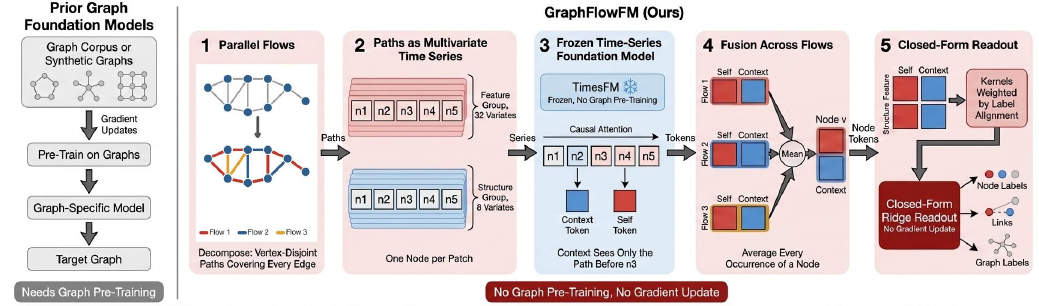}
  \caption{Prior graph foundation models (left) learn graph knowledge from real or synthetic graphs
  before they are applied to a target graph. \ours{} (right) decomposes the target graph into
  parallel flows, vertex-disjoint paths that jointly cover every edge (1), writes each path as a
  multivariate time series with one node per patch (2), and encodes it with a frozen time-series
  foundation model whose causal attention gives each node a self token and a context token (3).
  Since a node lies on paths of several flows, its tokens are averaged over all of its occurrences
  (4), and a closed-form ridge readout predicts node labels, links, or graph labels (5). The node
  features and the structural descriptors are encoded as two separate groups of variates, and no
  parameter is pre-trained on graphs or updated by gradient.}
  \label{fig:method}
\end{figure*}

\section{Related Work}
\label{sec:related}

\textbf{Graph foundation models.}
Graph foundation models differ in where their graph knowledge comes from \citep{wang2025gfmsurvey}.
Multi-domain pre-training learns it from collections of real graphs.
OFA \citep{liu2024ofa} and GOFA \citep{kong2025gofa} unify graphs through text attributes, and
GFT \citep{wang2024gft} and GIT \citep{wang2025git} through computation trees.
GCOPE \citep{zhao2024gcope}, SAMGPT \citep{yu2025samgpt}, MDGFM \citep{wang2025mdgfm}, and
RiemannGFM \citep{sun2025riemanngfm} align structures or geometries across domains, and R-GFM
\citep{liu2026rgfm} and GAUGE \citep{sun2026gauge} extend this line with Riemannian
graph-of-graphs and learned manifold geometry.
These models also address link prediction and graph classification, but every task still requires
pre-training on graphs and usually fine-tuning on the target graph.
Fully inductive models learn it from one graph.
GraphAny \citep{zhao2025graphany} combines closed-form linear GNNs with an attention module trained
on one graph.
TS-GNN \citep{finkelshtein2025equivariance} builds layers equivariant to node, label, and feature
permutations, and RGVT \citep{zhao2026viewspace} maps arbitrary graphs into a shared view space.
Prior-fitted models learn it from generated data.
NodePFN \citep{choi2026nodepfn} and GraphPFN \citep{eremeev2026graphpfn} pre-train on synthetic
graphs, and Node4All \citep{node4all2026} trains a channel graph transformer with self-supervision
on synthetic graphs.
G2T-FM \citep{eremeev2025g2tfm} and TAG \citep{hayler2025tag} feed neighbor-aggregated node
features to tabular foundation models \citep{hollmann2025tabpfn}.
LLM-BP \citep{wang2025llmbp} applies large language models to text-attributed graphs.
A recent re-evaluation finds that only the prior-fitted models outperform tuned GNNs
\citep{faireval2026}.
Unlike these models, \ours{} is not pre-trained on any graph, real or synthetic.
Its encoder is a time-series foundation model pre-trained only on time series, which reads each
graph through its parallel flows.

\textbf{Time-series foundation models and frozen transfer.}
TimesFM \citep{das2024timesfm}, Moirai \citep{woo2024moirai}, Chronos \citep{ansari2024chronos},
Time-MoE \citep{shi2025timemoe}, Moirai-MoE \citep{liu2025moiraimoe}, and Sundial
\citep{liu2025sundial} are pre-trained on large forecasting corpora and transfer without fine-tuning
across domains and frequencies.
Moirai, Moirai-MoE, and TimesFM-3 \citep{jain2026timesfm3} attend across variates, while the others
forecast each variate separately.
Their predictions are also better calibrated than those of baseline models
\citep{adler2026calibrated}, and in-context examples adapt them without gradient updates
\citep{faw2025icf}.
Frozen pretrained language transformers transfer to other modalities when only their input, output,
positional-embedding, and layer-normalization parameters are fitted \citep{lu2022fpt}, and tabular
foundation models have recently been turned into graph
foundation models \citep{eremeev2025g2tfm,hayler2025tag}.
\ours{} is, to our knowledge, the first to use a time-series foundation model as a graph foundation
model, and it keeps the model entirely frozen and fits only a closed-form readout.

\textbf{Sequence views of graphs and heterophily.}
GFCN \citep{ji2020gfcn} introduced parallel flows and bounded their number by the maximum degree,
then applied a one-dimensional convolutional network along the paths, trained on each graph.
\ours{} keeps the decomposition and replaces the trained convolution with a frozen time-series
foundation model.
Propagation-based models such as APPNP \citep{gasteiger2019appnp} and GPR-GNN \citep{chien2021gprgnn}
average neighbor features with fixed or learned weights.
On heterophilous graphs, such as the Roman-empire and Amazon-ratings benchmarks of
\citep{platonov2023critical}, this averaging mixes the features of nodes from different classes.
Ordered paths keep the position of each neighbor in the sequence, so \ours{} does not need to
average the neighbors of a node.
Its readout averages over neighbors only where this raises validation accuracy, and on Roman-empire
it mostly does not (\cref{sec:ana-flows}).

\section{Method}
\label{sec:method}

\ifdefined\methodfigearly\else\begin{figure*}[t]
  \centering
  \includegraphics[width=\textwidth]{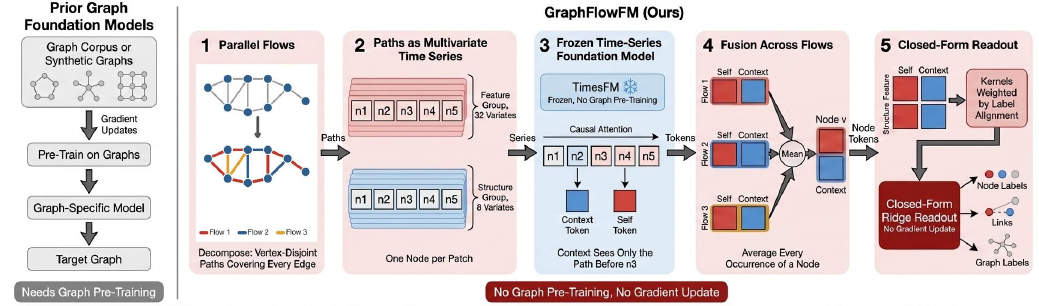}
  \caption{Prior graph foundation models (left) learn graph knowledge from real or synthetic graphs
  before they are applied to a target graph. \ours{} (right) decomposes the target graph into
  parallel flows, vertex-disjoint paths that jointly cover every edge (1), writes each path as a
  multivariate time series with one node per patch (2), and encodes it with a frozen time-series
  foundation model whose causal attention gives each node a self token and a context token (3).
  Since a node lies on paths of several flows, its tokens are averaged over all of its occurrences
  (4), and a closed-form ridge readout predicts node labels, links, or graph labels (5). The node
  features and the structural descriptors are encoded as two separate groups of variates, and no
  parameter is pre-trained on graphs or updated by gradient.}
  \label{fig:method}
\end{figure*}
\fi

\cref{fig:method} provides an overview of the \ours{} model. The remainder of this section is organized as follows: \cref{sec:setting} defines the problem setting, \cref{sec:flows} introduces the decomposition into parallel flows, and \cref{sec:encode} describes path encoding via a frozen time-series foundation model. We then detail the fusion of tokens across flows (\cref{sec:fusion}) and the closed-form readout (\cref{sec:readout}), followed by theoretical discussions in \cref{app:theory}.

\subsection{Setting}
\label{sec:setting}

We consider undirected graphs. Such a graph $G=(V,E)$ with $n=|V|$ nodes has adjacency matrix $A$, and node features $X\in\mathbb{R}^{n\times e}$, where $x_{vj}$ denotes feature $j$ of node $v$. The feature dimension $e$ and feature semantics may vary across graphs, and node features may be absent. Let $d_v$ denote the degree of node $v$ and $D=\operatorname{diag}(d_v:v\in V)$ the degree matrix.

We consider three prediction tasks, each formulated as assigning a label from $C$ classes to a prediction target. For node classification, each target is a node and its label is the node's class. For link prediction, each target is a node pair and its binary label indicates whether the pair forms an edge. For graph classification, each target is a graph in a collection, and its label describes a graph-level property, such as a molecule's biological activity. Given a labeled training set $L$ of targets and a labeled validation set, the goal is to predict the labels of test targets.

Our central idea is to represent a graph as a collection of ordered sequences by partitioning its edges into edge-disjoint paths. Each path preserves adjacency along the sequence, and the paths collectively cover every edge, while allowing a node to appear in multiple paths. We treat node features and structural descriptors along each path as multivariate time series and encode them using a frozen pretrained time-series foundation model. Parallel flows thus provide an interface between graph-structured data and pretrained sequence models.

Our model, \ours{}, uses the same frozen time-series foundation model across all datasets and tasks, without graph-specific pretraining. We also use a common set of hyperparameter settings throughout. Dataset-specific adaptation is confined to the downstream readout, which is fitted using the training labels and selected using the validation labels. No model parameter is updated by gradient descent.

\subsection{Parallel Flows}
\label{sec:flows}

\begin{definition}[Parallel flows\namecite{ji2020gfcn}]
A path is a connected acyclic subgraph in which every node has degree at most two.
A parallel flow is a collection of pairwise vertex-disjoint paths.
A collection of parallel flows covers $G$ if every edge of $G$ belongs to a path in at least one flow.
\end{definition}

Each path can be represented as an ordered sequence of nodes.
A parallel-flow cover therefore represents all edges of $G$ through sequences, while allowing paths in different flows to share nodes.
The following result bounds the number of flows required solely in terms of the maximum degree.

\begin{proposition}[{Number of flows\namecite[Theorem 1 and Corollary 1]{ji2020gfcn}}]
\label{prop:bound}
Let $d_{\max}$ be the maximum degree of $G$ and $h=\lfloor(d_{\max}+1)/2\rfloor$.
If $G$ is a tree, the minimum number of parallel flows needed to cover $G$ is $h$.
For a general graph, possibly disconnected, at most $h(h+1)$ flows suffice.
\end{proposition}

To understand the tree case, pair incident edges at each node and follow these pairs to form edge-disjoint paths.
A node of degree $d_v$ belongs to at most $\lceil d_v/2\rceil$ such paths.
These paths can be assigned to $h$ flows so that paths sharing a node belong to different flows.
Conversely, each flow covers at most two edges incident to any node, so the maximum-degree node requires at least $\lceil d_{\max}/2\rceil=h$ flows.
For a general graph, repeatedly extract and remove a spanning forest.
Each extraction reduces the maximum degree of the remaining graph by at least one, since every non-isolated node contributes at least one incident edge to the forest.
Applying the tree bound to each forest gives
\[
\sum_{\delta=1}^{d_{\max}}
\left\lfloor\frac{\delta+1}{2}\right\rfloor
\leq h(h+1).
\]

Following this forest-based construction, \ours{} computes a parallel-flow decomposition using \cref{alg:flows}.
At each iteration, we extract a depth-first search (DFS) forest from the remaining edges and remove its edges.
We partition each forest into edge-disjoint paths by pairing incident tree edges at each node.
The resulting paths are grouped into flows by greedy coloring: paths that share a node must receive different colors, and each color defines a flow.
The process continues until every edge has been assigned to a path.

The DFS uses a fixed traversal order: roots are considered in decreasing order of degree, and neighbors are visited in increasing order of degree.
This ordering favors exploration along low-degree chains.
Together with a fixed rule for breaking degree ties, it makes the traversal deterministic.

Within each forest, a node belongs to at most $\lceil d_v/2\rceil$ paths, and it may occur in paths from multiple forests.
A node can therefore appear in several sequences, with different neighbors preceding or following it.
We encode every path in both directions to capture context from either end.
Nodes not included in any path, including isolated nodes, are represented by singleton sequences.
Paths exceeding the encoder's context length are divided into windows of 480 nodes, with an overlap of 64 nodes between consecutive windows.

\begin{algorithm}[t]
  \caption{Parallel flow decomposition}
  \label{alg:flows}
  \begin{algorithmic}[1]
    \STATE \textbf{Input:} Graph $G=(V,E)$
    \STATE Initialize the uncovered edges $R\leftarrow E$ and the paths $\mathcal{P}\leftarrow\emptyset$
    \WHILE{$R\neq\emptyset$}
      \STATE Build a DFS forest $\mathcal{T}$ of $(V,R)$, with roots by decreasing degree and neighbors by increasing degree
      \STATE Pair the edges of $\mathcal{T}$ at each node, and follow the pairs into edge-disjoint paths
      \STATE Add the paths to $\mathcal{P}$, and $R\leftarrow R\setminus E(\mathcal{T})$
    \ENDWHILE
    \STATE Group $\mathcal{P}$ into flows by greedy vertex-disjoint coloring
    \STATE \textbf{Output:} Parallel flows that cover $E$
  \end{algorithmic}
\end{algorithm}

\subsection{Paths as Multivariate Time Series}
\label{sec:encode}

\textbf{Two complementary views of each node.}
To turn a path into a multivariate time series, we treat its nodes as successive positions and their attributes as signal channels, or \emph{variates}. We construct two complementary groups of signals: a \emph{feature group} that describes what each node contains, and a \emph{structure group} that describes its local connectivity. We encode the two groups separately, keeping each within the encoder's limit of 32 variates.

For the feature group, we normalize the available node features and project them onto their leading singular-vector directions, obtaining 32 channels. This label-independent transformation accommodates differences in feature dimension and semantics across graphs. We denote the resulting signal matrix by $S_{\mathrm{f}}\in\mathbb{R}^{n\times p_{\mathrm{f}}}$, where $p_{\mathrm{f}}=32$.

For the structure group, we use $8$ channels: the log-transformed degree $\log(1+d_v)$ and the probabilities that a random walk starting at $v$ returns to $v$ after one through seven steps. The return probability after $r$ steps is $((D^{-1}A)^r)_{vv}$. Together, these signals characterize local connectivity at several walk lengths and remain available when node features are absent. We collect them in $S_{\mathrm{s}}\in\mathbb{R}^{n\times p_{\mathrm{s}}}$, where $p_{\mathrm{s}}=8$. Each column of both signal matrices is standardized. The feature preprocessing rules are specified in \cref{app:impl}.

\textbf{From paths to encoder patches.}
We encode these signals using the frozen multivariate time-series foundation model TimesFM-3 \citep{jain2026timesfm3}, the successor of TimesFM \citep{das2024timesfm}. The path determines the sequence order; no temporal interpretation of the graph is required. The encoder processes sequences in patches of $P=32$ time points, so we assign one patch to each node. Within that patch, we repeat each signal value across all $P$ positions. Thus, a path $q=(v_1,\ldots,v_{|q|})$ becomes a multivariate series of length $|q|P$, with one channel for each column of the chosen signal matrix.

We disable the encoder's built-in input normalization, replacing it with the identity transformation. This normalization rescales each input series using its own running mean and scale, in the spirit of reversible instance normalization \citep{kim2022revin}, and can obscure signal levels needed for node prediction. In a reconstruction check on Cora, ridge regression from the hidden states to the node's input signal achieves $R^2=0.03$ with this normalization and $R^2=0.99$ without it. We therefore retain the globally standardized signals without further input normalization.

At each node position, the encoder produces a hidden state for every variate. For signal group $g\in\{\mathrm{f},\mathrm{s}\}$, we concatenate these states into $H_i^{(q)}\in\mathbb{R}^{p_gd}$, where $d=1280$ is the hidden width. Causal attention gives two useful views of node $v_i$: $H_i^{(q)}$ summarizes the path prefix including $v_i$, whereas $H_{i-1}^{(q)}$ summarizes only the preceding nodes. These provide the basis for the \emph{self and context} representations introduced next.

\subsection{Fusion Across Flows}
\label{sec:fusion}

\textbf{Combining multiple views of a node.}
A node can appear in several sequences, each exposing a different ordered context. We propose to average its hidden states across these occurrences to obtain a single node representation, while also retaining a complementary representation by averaging the hidden states immediately preceding the node. These two representations capture the path context after and before the encoder reads the node's signal, respectively. Keeping them separate allows the downstream readout to determine their relative usefulness for prediction. Each occurrence presents the encoder with a different ordered context. \ours{} combines these views into two representations per signal group: a self token, which includes the node's own signal, and a context token, which summarizes the sequence preceding it.

Specifically, for an occurrence of node $v$ at position $i$ in sequence $q$, the hidden state $H_i^{(q)}$ includes both $v$ and its preceding path context. The previous hidden state, $H_{i-1}^{(q)}$, summarizes that context before the encoder reads $v$'s patch. We average these states separately across occurrences, allowing each node to gather information from the different paths through it.

Formally, let $\mathcal{O}(v)$ contain all occurrences $(q,i)$ of $v$ across flows, paths, windows, and traversal directions. Let $\mathcal{O}^{-}(v)$ contain those occurrences with a predecessor, namely $i>1$. For each signal group $g\in\{\mathrm{f},\mathrm{s}\}$, we define
\begin{align}
z^{\mathrm{self}}_{g,v}
&=\frac{1}{|\mathcal{O}(v)|}
  \sum_{(q,i)\in\mathcal{O}(v)} H_i^{(q)},
\label{eq:tokens}\\
z^{\mathrm{ctx}}_{g,v}
&=\frac{1}{|\mathcal{O}^{-}(v)|}
  \sum_{(q,i)\in\mathcal{O}^{-}(v)} H_{i-1}^{(q)}.
\nonumber
\end{align}
Here, the hidden states are computed from signal group $g$. We set the context token to zero when $\mathcal{O}^{-}(v)$ is empty.

Because the paths collectively cover every edge and are encoded in both directions, each neighbor can contribute to the context preceding $v$ in at least one sequence. Averaging brings these path-based views together, while keeping the self and context information in separate tokens. The resulting representations summarize the contexts exposed by the chosen decomposition; they need not be invariant to the decomposition itself.

Stacking the tokens as rows, one per node, gives four representation blocks: $Z^{\mathrm{self}}_{\mathrm{f}}$, $Z^{\mathrm{ctx}}_{\mathrm{f}}$, $Z^{\mathrm{self}}_{\mathrm{s}}$, and $Z^{\mathrm{ctx}}_{\mathrm{s}}$. This fusion step introduces no learned parameters.

\subsection{Closed-Form Readout}
\label{sec:readout}
The frozen encoder provides several views of each node, but their usefulness depends on the task. Our readout combines these views with different amounts of graph propagation and uses labeled examples to identify useful similarities between prediction targets. A closed-form predictor and validation-selected ensemble provide this adaptation without updating the encoder.

\textbf{Blocks and propagation.}
The readout combines five complementary views of each node: the four token blocks $Z^{\mathrm{self}}_{\mathrm{f}}$, $Z^{\mathrm{ctx}}_{\mathrm{f}}$, $Z^{\mathrm{self}}_{\mathrm{s}}$, and $Z^{\mathrm{ctx}}_{\mathrm{s}}$, together with the input feature signals $S_{\mathrm{f}}$. Each block is standardized columnwise. To supplement path context with graph connectivity, we consider no propagation, one- and two-step propagation, and personalized PageRank propagation. With $\hat A=\tilde D^{-1/2}(A+I)\tilde D^{-1/2}$, where $\tilde D$ is the degree matrix of $A+I$, these are $\Phi_k(Z)=\hat A^kZ$ for $k\in\{0,1,2\}$ and
\begin{equation}
\Phi_{\mathrm{ppr}}(Z)
=\sum_{\tau=0}^{T-1}\alpha(1-\alpha)^\tau\hat A^\tau Z
 +(1-\alpha)^T\hat A^TZ,
\label{eq:ppr}
\end{equation}
with restart probability $\alpha=0.1$ and $T=10$ \citep{gasteiger2019appnp}. Keeping all four versions allows the readout to accommodate different amounts of neighborhood mixing.

\textbf{Target representations.}
We adapt the node representations to each task through a simple aggregation. For a block $Z$ and propagation $\phi\in\{\Phi_0,\Phi_1,\Phi_2,\Phi_{\mathrm{ppr}}\}$, define
\begin{equation}
\psi(Z)_a=
\begin{cases}
\phi(Z)_v, & a=v,\\
\phi(Z)_u\odot\phi(Z)_v, & a=(u,v),\\
\displaystyle\frac{1}{|V_a|}\sum_{v\in V_a}\phi(Z)_v,
& a\text{ is a graph}.
\end{cases}
\label{eq:target}
\end{equation}
Thus, node classification uses individual rows, link prediction uses an order-independent elementwise product, and graph classification uses mean pooling. We suppress the dependence on $\phi$ and write $\psi(Z)_L$ for the rows corresponding to training targets.

\textbf{Kernels and block combinations.}
We compare targets using either inner-product similarity or Gaussian distance similarity:
\begin{align*}
&K^{\mathrm{lin}}(Z)=\psi(Z)\psi(Z)_L^\top,\\
&K^{\mathrm{rbf}}(Z)_{ab}
=\exp\!\left(-\frac{\|\psi(Z)_a-\psi(Z)_b\|^2}{s^2}\right).
\end{align*}
The linear kernel is normalized by its mean training diagonal, and $s^2$ is the median squared distance between training targets.

Because feature and structural information may differ in usefulness, we consider six kernel combinations: all four token blocks; feature self only; feature self plus context; structure self only; structure self plus context; and a weighted sum over all five blocks. For the weighted sum, each weight is the nonnegative part of the centered kernel-target alignment with the training label kernel \citep{cortes2012alignment}, normalized across blocks. Each combination uses either linear or Gaussian kernels throughout. Without node features, only the two structure-based combinations are used.

\textbf{Closed-form readout.}
We fit a kernel ridge predictor for each combination, adapting to the labels without updating the encoder. For a combined kernel $K$, the class scores are
\begin{equation}
F=K(K_{LL}+\lambda\mu I)^{-1}
(Y_L-\mathbf{1}\pi^\top)+\mathbf{1}\pi^\top,
\label{eq:ridge}
\end{equation}
where $K_{LL}$ is the training kernel, $Y_L$ contains one-hot training labels, $\pi$ contains training class proportions, $\mu$ is the mean diagonal of $K_{LL}$, and $\lambda\in\{0.1,1,10\}$. The class proportions provide a baseline prediction, while the dual formulation requires solving a $|L|\times|L|$ system.

Scores become probabilities through the rowwise transformation $Q=\mathrm{softmax}(3F/\sigma_F)$, where $\sigma_F$ is the standard deviation of the entries of $F$. For node classification, we also retain a smoothed prediction obtained from ten iterations of $Q\leftarrow(1-\alpha)\hat A Q+\alpha Q_0$, starting from $Q_0=Q$, followed by row normalization.

\textbf{Ensemble selection.}
The six combinations, two kernels, four propagations, and three ridge strengths yield 144 candidates, or 288 for node classification with smoothing; fewer are used without node features. Greedy ensemble selection with replacement \citep{caruana2004ensemble} selects one candidate at each of 30 steps to maximize the validation score of the running average: accuracy for node classification and ROC-AUC otherwise. The final prediction averages the selected candidates. The candidate settings and selection procedure are shared across datasets, while validation labels determine the ensemble.

\section{Theoretical Analysis: Consistency of Single-Decomposition Empirical Risk}
\label{app:theory}

A graph generally admits many parallel-flow decompositions, and the output of
\cref{alg:flows} can depend on its vertex order (\cref{app:relabel}).
This raises a natural question: must we average over multiple decompositions
of every graph to obtain a meaningful population-level evaluation?
We show that, for a fixed predictor and independently sampled graphs, one
decomposition per graph consistently estimates the risk averaged over a
specified decomposition distribution.
The key idea is to average jointly over graph examples and decomposition
choices, rather than exhaustively over the decompositions of each graph.
This result provides a statistical rationale for single-decomposition
processing; its broader relevance to the transferable design of \ours{}
is discussed below.

\textbf{Setting.}
Let $(G,Y)\sim\mathcal{P}$ denote a graph, including its observed attributes,
and its prediction target or collection of targets.
For node-level tasks, the loss of a graph can be an average over its target nodes;
the graph remains the sampling unit.
Let $\mathfrak{D}(G)$ be the set of parallel-flow decompositions of $G$, and
let $\mu_G$ be a probability distribution on this set that depends measurably
on $G$.
For \cref{alg:flows} applied after a uniformly random vertex relabeling,
$\mu_G$ is the distribution of the resulting decomposition, mapped back to
the original vertices.
For the deterministic procedure using the original vertex identifiers,
$\mu_G=\delta_{\mathcal{A}(G)}$ is the point mass at its output
$\mathcal{A}(G)$.
Thus, the analysis covers both randomized and deterministic procedures,
but always estimates the risk induced by the specified procedure.

Write the predictor as
$f_\theta(G,D)=h_\theta(F_\phi(G,D))$,
where $F_\phi$ includes signal construction, the frozen TimesFM encoder,
and token fusion and propagation, while $h_\theta$ denotes a fixed readout,
including its ridge coefficients and selected ensemble.
For a measurable loss $\ell$, define the graph loss, its decomposition
average, and the population risk as
\begin{align}
 L_\theta(G,Y,D)
 &=\ell\big(f_\theta(G,D),Y\big),
 \label{eq:sd-loss}\\
 m_\theta(G,Y)
 &=\int_{\mathfrak{D}(G)}L_\theta(G,Y,D)\,\mu_G(dD),
 \label{eq:sd-conditional}\\
 R(\theta)
 &=\mathbb{E}_{(G,Y)\sim\mathcal{P}}[m_\theta(G,Y)].
 \label{eq:sd-risk}
\end{align}
Here $R(\theta)$ is the expected loss when each graph is processed using one
decomposition drawn from $\mu_G$; it is not the loss of an ensemble that
averages predictions over decompositions.
Given one decomposition $D_i$ for each of $N$ graph examples, the empirical
risk is
\begin{equation}
 \widehat R_N(\theta)
 =\frac1N\sum_{i=1}^N L_\theta(G_i,Y_i,D_i).
 \label{eq:sd-empirical}
\end{equation}
We assume:
\begin{itemize}
 \item[(A1)] The examples $(G_i,Y_i)_{i\geq1}$ are independent and
 identically distributed with law $\mathcal{P}$.
 \item[(A2)] Conditional on all graph examples $(G_i,Y_i)$, the
 decompositions are independent, with $D_i\sim\mu_{G_i}$.
 The same label-independent decomposition procedure is used for every graph.
 \item[(A3)] The predictor parameter $\theta$ is fixed independently of
 the sampled examples and decompositions, and
 $\mathbb{E}|L_\theta(G,Y,D)|<\infty$ under
 $\mathcal{P}(dG,dY)\mu_G(dD)$.
\end{itemize}

\begin{proposition}[Consistency of single-decomposition empirical risk]
\label{prop:single}
Under (A1)--(A3),
$\mathbb{E}[\widehat R_N(\theta)]=R(\theta)$ and
$\widehat R_N(\theta)\to R(\theta)$ almost surely as $N\to\infty$.
If, in addition, $0\leq L_\theta\leq M$ almost surely for some $M>0$, then,
for every $\varepsilon>0$,
\begin{equation}
 \Pr\!\left(\left|\widehat R_N(\theta)-R(\theta)\right|
 \geq\varepsilon\right)
 \leq2\exp\!\left(-\frac{2N\varepsilon^2}{M^2}\right).
 \label{eq:sd-hoeffding}
\end{equation}
Consequently, with probability at least $1-\delta$, where $0<\delta<1$,
\[
 \left|\widehat R_N(\theta)-R(\theta)\right|
 \leq M\sqrt{\frac{\log(2/\delta)}{2N}}.
\]
\end{proposition}

\begin{proof}
Let $Z_i=(G_i,Y_i,D_i)$ and define the joint law
$Q(dG,dY,dD)=\mathcal{P}(dG,dY)\mu_G(dD)$.
We first verify that the $Z_i$ are independent and identically distributed
with law $Q$.
For measurable sets $B_1,\ldots,B_N$, let
$q_i(G,Y)=\int\mathbf{1}_{B_i}(G,Y,D)\mu_G(dD)$.
Conditional independence in (A2), followed by independence in (A1), gives
\begin{align*}
 \Pr(Z_i\in B_i\text{ for all }i)
 &=\mathbb{E}\!\left[\prod_{i=1}^N q_i(G_i,Y_i)\right]\\
 &=\prod_{i=1}^N\mathbb{E}[q_i(G_i,Y_i)]
 =\prod_{i=1}^N Q(B_i).
\end{align*}
For fixed $\theta$, the losses $X_i=L_\theta(G_i,Y_i,D_i)$ are therefore
also i.i.d., and (A3) ensures integrability.
By (A2) and the tower property,
\[
 \mathbb{E}[X_i\mid G_i,Y_i]=m_\theta(G_i,Y_i),
 \qquad \mathbb{E}[X_i]=R(\theta).
\]
Linearity of expectation proves unbiasedness, and the strong law of large
numbers gives
$N^{-1}\sum_{i=1}^N X_i\to\mathbb{E}[X_1]=R(\theta)$ almost surely.
Neither boundedness nor a finite second moment is required for these
conclusions.

If $0\leq X_i\leq M$, Hoeffding's inequality bounds each tail of
$\sum_{i=1}^N(X_i-\mathbb{E}X_i)$ at $N\varepsilon$ by
$\exp(-2N\varepsilon^2/M^2)$.
Adding the two tail bounds proves \eqref{eq:sd-hoeffding}.
Setting its right-hand side to $\delta$ yields the confidence bound.
\end{proof}

Thus, $N\geq M^2\log(2/\delta)/(2\varepsilon^2)$ independent graph
examples suffice for the stated accuracy and confidence.
The bounded-loss condition holds with $M=1$ for graph-level classification
error or an average of node-level classification errors.

\textbf{Why one decomposition can suffice.}
When the loss has finite variance, the law of total variance gives
\begin{equation}
 \operatorname{Var}(X_i)
 =\operatorname{Var}(m_\theta(G_i,Y_i))
 +\mathbb{E}\!\left[
   \operatorname{Var}(X_i\mid G_i,Y_i)\right].
 \label{eq:sd-variance}
\end{equation}
The two terms reflect variation between graph examples and variation
between decompositions of the same graph.
Since $\operatorname{Var}(\widehat R_N)=\operatorname{Var}(X_i)/N$,
averaging over more independent graphs reduces both contributions.
Averaging the losses of $K$ conditionally independent decompositions per
graph reduces only the second contribution by a further factor of $K$.
Repeated decompositions can therefore improve precision, but are not
necessary for consistency of the population-risk estimate.

\textbf{Relevance and scope.}
The proposition provides a population-level rationale for processing each
graph with a single decomposition.
More broadly, this perspective is consistent with the \emph{transferable} design
of \ours{}, which reuses a frozen encoder across graphs and tasks.
It suggests that exhaustive averaging over decompositions need not be
necessary for effective prediction across graph examples.
This broader interpretation motivates the approach, while its practical
relevance is assessed empirically; it is not an additional guarantee of
the proposition.

The theoretical analysis establishes consistency of population-risk estimation for a fixed predictor evaluated on independent graph examples. This includes predictors fitted on independent training data and cross-domain generalization tasks, alluded to by foundation models, across graphs, with each graph contributing an aggregate loss. It also provides a starting point for further analysis of the complete learning pipeline. Promising \emph{future directions} include generalization guarantees for data-dependent readout fitting and selection, and extensions that account for dependence among node or edge samples within a graph. Complementary theoretical and empirical studies could characterize stability under vertex relabeling and decomposition choices, and clarify how time-series pretraining contributes to effective graph representations.

\section{Experiments}
\label{sec:experiments}

We evaluate \ours{} on three tasks: node classification (\cref{sec:exp-node}), link prediction (\cref{sec:exp-link}), and graph classification (\cref{sec:exp-graph}). Dataset details and evaluation protocols are provided in \cref{app:protocols}. For each task, we compare \ours{} with methods trained on graph data. We then examine the contributions of its components in \cref{sec:analysis}.

\textbf{Common setup.}
All experiments use the same public TimesFM-3 checkpoint with its parameters frozen. The hyperparameter settings and readout candidate library described in \cref{sec:method} are shared across all datasets and tasks (\cref{tab:hparams}, \cref{app:impl}). Readout candidates are fitted using training labels, and the ensemble is selected separately for each dataset using validation labels (\cref{sec:readout}). All experiments run on a single NVIDIA RTX 4090D GPU with 24\,GB of memory.

\subsection{Node Classification}
\label{sec:exp-node}

\textbf{Datasets and protocol.}
We evaluate on the node classification datasets of GraphAny \citep{zhao2025graphany}, the largest
set of datasets on which recent graph foundation models have publicly reported results under a
common evaluation protocol.
The main comparison uses 21 datasets from nine domains: citation, co-purchase, co-authorship, web,
text, social, crowdsourcing, co-occurrence, and a synthetic grid.
We group them by edge homophily, above or below 0.5, into 14 homophilous and 7 heterophilous
datasets.
\cref{tab:data} in \cref{app:data} lists every dataset with its source and statistics.
We use the official data splits where available. Otherwise, we sample 20 labeled nodes per class for training and divide the remaining nodes equally between validation and test sets using class-stratified sampling. We report mean test accuracy over five random seeds.

\textbf{Baselines.}
We compare with graph foundation models published at top venues in 2025 and 2026 that report
results under this protocol, GraphAny \citep{zhao2025graphany}, TS-Mean and TS-GAT
\citep{finkelshtein2025equivariance}, NodePFN \citep{choi2026nodepfn}, and RGVT
\citep{zhao2026viewspace}.
All baseline methods are trained on graph data: GraphAny, TS-GNN, and RGVT use real graphs, whereas NodePFN uses synthetic graphs. RGVT additionally fits a lightweight predictor on each target dataset. We use the results reported in the original papers for GraphAny and TS-GNN, and those in Tables 16 and 17 of \citet{zhao2026viewspace} for NodePFN and RGVT. When multiple variants are available, we follow each paper's main configuration, using GraphAny trained on Wisconsin and RGVT with an MLP head. \cref{app:full} reports results for all variants and includes TAG \citep{hayler2025tag}, which has not yet been formally published.

\begin{table}[t]
\centering
\caption{Node classification accuracy (\%) under the GraphAny protocol on 21 datasets, five seeds. Baselines are released numbers of graph foundation models from 2025 and 2026, and every baseline is trained on graph data. n/r means not reported. Rank is averaged over the datasets a method reports. Best in bold, second underlined. Standard deviations and further variants are in \Cref{app:full}.}
\label{tab:main}
\scriptsize
\setlength{\tabcolsep}{1.2pt}
\begin{tabular}{lcccccc}
\toprule
Dataset & GraphAny & TS-Mean & TS-GAT & NodePFN & RGVT & \textbf{Ours} \\
\midrule
\multicolumn{7}{l}{\emph{Homophilous}} \\
Cora & 77.8 & n/r & n/r & \underline{82.1} & 81.7 & \textbf{83.1} \\
CiteSeer & 67.5 & 68.7 & 68.7 & 67.3 & \underline{70.1} & \textbf{70.4} \\
PubMed & 77.5 & 75.0 & 75.5 & 78.0 & \underline{78.2} & \textbf{79.3} \\
Photo & 90.2 & 90.2 & 89.7 & 90.5 & \underline{92.3} & \textbf{92.5} \\
CS & 90.8 & 90.9 & 91.1 & 91.5 & \underline{92.3} & \textbf{93.0} \\
Computers & 82.0 & 81.4 & 80.2 & 81.4 & \underline{84.1} & \textbf{86.6} \\
Physics & 92.5 & 92.6 & 92.2 & \textbf{93.4} & \underline{93.2} & 92.9 \\
DBLP & 70.1 & 66.4 & 64.3 & 74.7 & \textbf{78.1} & \underline{77.2} \\
FullCora & 56.3 & 53.6 & 52.8 & n/r & \textbf{63.3} & \underline{60.8} \\
Wiki & 61.1 & 69.9 & \underline{72.5} & n/r & 72.3 & \textbf{75.2} \\
WikiCS & 73.8 & 74.2 & 73.2 & 76.0 & \underline{78.9} & \textbf{79.9} \\
Minesweeper & 80.1 & 80.7 & \underline{80.7} & 80.7 & 80.0 & \textbf{82.9} \\
Tolokers & 78.2 & 78.1 & 78.0 & \underline{78.6} & 78.3 & \textbf{80.5} \\
Questions & 97.1 & 97.0 & 97.0 & 97.0 & \underline{97.1} & \textbf{97.3} \\
\midrule
\multicolumn{7}{l}{\emph{Heterophilous}} \\
Roman-empire & 64.1 & 66.4 & 67.8 & n/r & \underline{70.5} & \textbf{80.6} \\
Amazon-ratings & 42.6 & 42.3 & 41.9 & 44.7 & \underline{45.6} & \textbf{50.7} \\
BlogCatalog & 77.7 & 76.3 & 78.4 & n/r & \underline{81.4} & \textbf{90.3} \\
Cornell & 66.5 & 68.7 & 71.3 & 71.9 & \textbf{74.0} & \underline{73.0} \\
Texas & 73.5 & 73.5 & 74.0 & 76.2 & \underline{78.9} & \textbf{84.3} \\
Wisconsin & 71.8 & 61.2 & 65.9 & \underline{79.2} & 78.0 & \textbf{81.2} \\
Actor & 29.5 & 28.1 & 28.8 & \underline{33.0} & 32.7 & \textbf{33.4} \\
\midrule
Avg.\ rank, homophilous & 4.29 & 4.38 & 4.69 & 3.17 & 2.36 & \textbf{1.29} \\
Avg.\ rank, heterophilous & 4.57 & 5.14 & 4.29 & 2.60 & 2.14 & \textbf{1.14} \\
Avg.\ rank, all 21 & 4.38 & 4.65 & 4.55 & 3.00 & 2.29 & \textbf{1.24} \\
Datasets ranked first & 0/21 & 0/20 & 0/20 & 1/17 & 3/21 & \textbf{17}/21 \\
\bottomrule
\end{tabular}
\end{table}

\textbf{Main results.}
\cref{tab:main} shows that \ours{} attains the best average rank at 1.24, ahead of RGVT at 2.29 and
NodePFN at 3.00, and ranks first on 17 of the 21 datasets.
It ranks first on 11 of 14 homophilous datasets (average rank 1.29) and on 6 of 7 heterophilous
datasets (1.14, against 2.14 for RGVT), where the margins are larger.
\ours{} is ahead of the strongest competitor by 10.0 points on Roman-empire, 8.9 on BlogCatalog,
5.4 on Texas, and 5.2 on Amazon-ratings.

\ours{} is behind on four datasets, by 0.5 points on Physics, by 0.9 on DBLP, by 1.1 on Cornell, a
web graph with 183 nodes, and by 2.4 on FullCora.
These gaps are computed from the unrounded accuracies in \cref{tab:full}.
A possible reason that \ours{} slightly underperforms on FullCora is that its 70 classes outnumber the 32 feature variates. With every reported variant and TAG added, \ours{} keeps the best average rank, 1.52 against 2.95
for RGVT (\cref{fig:teaser}), and \cref{tab:full} in \cref{app:results} reports all variants on all datasets.

\subsection{Link Prediction}
\label{sec:exp-link}

\begin{figure}[t]
  \centering
  \includegraphics[width=\columnwidth]{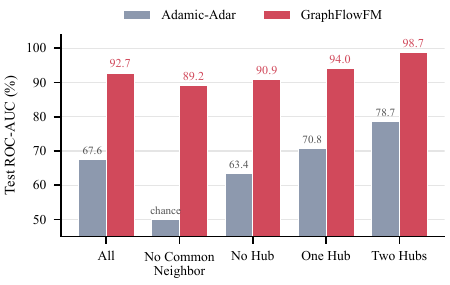}
  \caption{Link prediction on Cora, test ROC-AUC on groups of test pairs, against the Adamic-Adar
  heuristic, which scores a pair by its common neighbors weighted by inverse log degree. Hubs are the top 10\% of nodes by degree in the training graph, and
  the groups No Hub, One Hub, and Two Hubs contain the test pairs with zero, one, or two hub
  endpoints. No Common Neighbor contains the pairs whose two nodes share no neighbor in the training
  graph.}
  \label{fig:link}
\end{figure}

\textbf{Protocol and baselines.}
Following the standard protocol described by \citet{wang2025git}, we split the edges at
random into 70\% training, 15\% validation, and 15\% test edges with as many uniformly sampled
non-edges. For \ours{}, we compute flows and signals on the training edges only (\cref{app:link-protocol}).

\begin{table}[t]
\centering
\caption{Link prediction (ROC-AUC, \%). For each dataset, the best released result of a graph
foundation model that is pre-trained on graphs and fine-tuned on the target graph, and of a model
trained on the target graph alone. \ours{} makes no gradient update. \cref{tab:link} in \cref{app:link} lists every
baseline.}
\label{tab:link-short}
\scriptsize
\setlength{\tabcolsep}{1.8pt}
\begin{tabular}{lllc}
\toprule
Dataset & Best pre-trained GFM & Best per-dataset model & \ours{} \\
\midrule
Cora & 89.27 (R-GFM) & 87.34 (GCN) & \textbf{92.86} \\
CiteSeer & 90.88 (R-GFM) & \textbf{96.90} (DGI) & 93.81 \\
PubMed & 92.26 (OFA) & 91.16 (GCN) & \textbf{95.89} \\
AirUS & 86.17 (GCOPE) & 92.50 (DGI) & \textbf{95.78} \\
Amazon-ratings & 98.49 (GIT-S) & 98.05 (BGRL) & \textbf{99.70} \\
GitHub & 86.72 (OpenGraph) & 87.48 (GCN) & \textbf{91.69} \\
\bottomrule
\end{tabular}
\end{table}

\textbf{Results.}
\cref{tab:link-short} compares \ours{} with reported results of multiple baselines including R-GFM \citep{liu2026rgfm}, GAUGE
\citep{sun2026gauge}, and GIT \citep{wang2025git}.
\ours{} has the highest ROC-AUC on five of the six datasets and exceeds every compared pre-trained
model on all six, including GitHub and AirUS, which only use structures.

\textbf{Pairs without common neighbors.}
\cref{fig:link} examines where the advantage comes from on Cora.
Adamic-Adar \citep{adamic2003friends} is a classic link prediction heuristic that predicts a link
between two nodes from their common neighbors, each weighted by the inverse logarithm of its degree.
It uses no node features and gives the same score of zero to every pair without common neighbors.
Comparing with it on the same test pairs therefore tests whether \ours{} relies only on common
neighbors.
\ours{} reaches 92.7 ROC-AUC against 67.6.
On the 1,298 test pairs whose two nodes share no neighbor in the training graph, the heuristic is at
chance, while \ours{} still reaches 89.2.
The comparison shows that \ours{} predicts links even between nodes without common neighbors.
Its tokens encode the features of the nodes along the paths through each node, so two nodes with
similar path contexts receive similar tokens even when they share no neighbor.
\ours{} therefore predicts the links that heuristics based on common neighbors miss.
Because a path continues beyond the immediate neighbors of a node, the parallel flows carry this
wider context to the frozen time-series encoder, which has never been trained on a graph yet turns
it into tokens that predict links.

\subsection{Graph Classification}
\label{sec:exp-graph}

\begin{table}[t]
\centering
\caption{Graph classification (ROC-AUC, \%) on OGB molecule datasets with scaffold splits. Baselines are released numbers from \citet{wang2025git}, trained on the target data from scratch (Per-Dataset) or after graph pre-training (Pre-Train + FT). \ours{} makes no gradient update. Best bold, second underlined.}
\label{tab:graphcls}
\scriptsize
\setlength{\tabcolsep}{1.2pt}
\begin{tabular}{llcccccc}
\toprule
Method & Graph training & HIV & BBBP & BACE & ToxCast & Tox21 & Rank \\
\midrule
GIN & Per-Dataset & \underline{76.83} & 67.36 & 75.55 & 62.92 & 72.26 & 4.6 \\
BGRL & Per-Dataset & 72.18 & 67.40 & 73.75 & 62.52 & 72.97 & 5.4 \\
GraphMAE & Per-Dataset & 69.54 & 66.43 & 66.56 & 62.52 & \underline{74.13} & 5.4 \\
OFA & Pre-Train + FT & 76.48 & 65.79 & 77.88 & \underline{63.49} & 73.00 & 4.0 \\
GIT-G & Pre-Train + FT & 73.63 & 68.33 & \underline{79.28} & 63.00 & 73.81 & 3.2 \\
GIT-S & Pre-Train + FT & 74.75 & \textbf{68.72} & \textbf{81.10} & \textbf{63.63} & 73.78 & 2.2 \\
\midrule
\textbf{Ours} & \textbf{No Gradient Update} & \textbf{78.02} & \underline{68.59} & 75.87 & 61.95 & \textbf{74.46} & 3.0 \\
\bottomrule
\end{tabular}
\end{table}

\begin{figure}[t]
  \centering
  \includegraphics[width=\columnwidth]{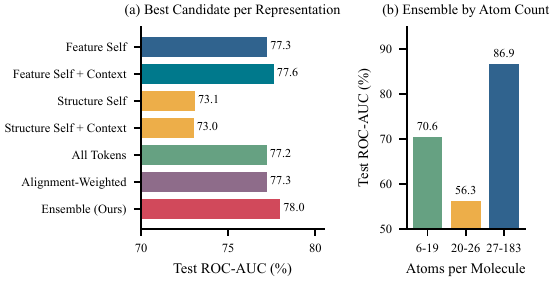}
  \caption{Graph classification on HIV. (a) Test ROC-AUC of the best single candidate of each
  representation, selected on the validation set, and of the ensemble. (b) Test ROC-AUC of the ensemble on the test molecules split into three equal groups by number of atoms.}
  \label{fig:graph}
\end{figure}

\begin{figure*}[t]
  \centering
  \includegraphics[width=\textwidth]{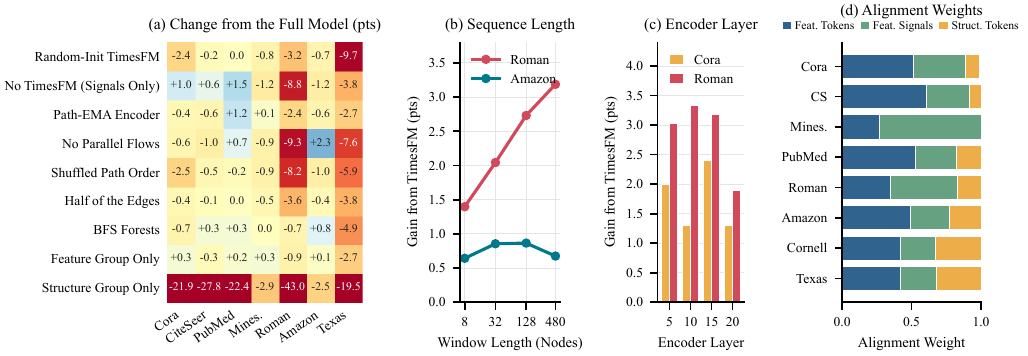}
  \caption{Node classification analyses. (a) Change in test accuracy from the full \ours{} when one
  component is changed. (b) Gain from TimesFM, the accuracy of the
  pretrained TimesFM minus that of a randomly initialized TimesFM of the same architecture, when
  every path is cut into windows of at most the given length. (c) The same gain when the tokens are read at different encoder layers. (d) Share of the
  alignment weight that the readout gives to the feature tokens, the feature signals, and the
  structure tokens. Exact numbers are in \cref{app:analyses} (\cref{fig:ablation-full,fig:length-layer,fig:align}).}
  \label{fig:node}
\end{figure*}

\textbf{Protocol and baselines.}
To our knowledge, no graph foundation model classifies graphs without training on graph data. Thus, on five molecule datasets of OGB with scaffold splits (\cref{app:graph}), \cref{tab:graphcls}
compares \ours{} with the fine-tuned results reported by \citet{wang2025git}.
Every baseline is trained on the target molecules, and OFA and GIT are also pre-trained on graphs.
We use one decomposition per molecule, and \cref{app:theory} discusses when this suffices for an
averaged loss.

\textbf{Results.}
\ours{} ranks first on HIV and Tox21 and second on average rank, behind GIT-S, which is pre-trained
on molecules and fine-tuned on each task.
Its average rank is 3.0, against 2.2 for GIT-S, 3.2 for GIT-G, and 4.6 for a GIN trained on each
task.
\ours{} thus competes with models trained on molecules, although its encoder has never seen one.

\textbf{Complementary feature and structure tokens.}
\cref{fig:graph}a examines HIV, where \ours{} ranks first, by evaluating the best single
candidate of each representation.
The feature tokens alone reach 77.3 to 77.6 ROC-AUC, and the structure tokens reach 73.1 without
any atom type.
Combining both groups reaches 78.0, above either group alone, and it is most accurate on molecules
with 27 or more atoms (\cref{fig:graph}b).
The experiment shows that both the feature information and the structure information are useful
for classifying molecules, and that they complement each other.
\ours{} encodes both groups and lets the readout combine them, so it outperforms either group
alone.
The structure tokens read only degree and random-walk signals along the parallel flows, and the
frozen time-series encoder, which has never seen a molecule, extracts useful information from both
groups.

\section{Analysis}
\label{sec:analysis}

This section asks where the accuracy of \ours{} comes from.
We separate what the pretrained encoder adds (\cref{sec:ana-encoder}) from what the parallel flows
add (\cref{sec:ana-flows}), measure the dependence on the vertex order (\cref{sec:ana-order}),
examine the two signal groups (\cref{sec:ana-groups}), and report the computational time (\cref{sec:exp-impl}).
The analyses use node classification, which has the most datasets, and each change keeps
everything else in \cref{sec:method} fixed.

\subsection{What the Pretrained Encoder Adds}
\label{sec:ana-encoder}

\textbf{Encoder variants.}
To separate the encoder from the readout, we replace only TimesFM and keep the signals, the flows,
the fusion, the propagation, the candidates, and the selection unchanged.
A randomly initialized TimesFM has the same architecture as the pretrained one.
Without TimesFM, both tokens of node $v$ in group $g$ are its row $(S_g)_v$, so the order along the
paths has no effect and the variant measures the readout on the signals alone.
The two path encoders compute $h_i$ at position $i$ of every sequence $q=(v_1,\dots,v_{|q|})$, the
mean of the signals up to $i$ for the path-mean encoder and
$h_i = \gamma h_{i-1} + (1-\gamma)(S_g)_{v_i}$ with $\gamma = 0.5$ for the path-EMA encoder.
As with TimesFM, the self token of a node is $h_i$ at the node, the context token is $h_{i-1}$, and
both are averaged over the occurrences of the node.
Comparing \ours{} with these variants tests whether its gains come from the pretrained weights of
TimesFM or only from ordering the signals along paths.

\textbf{The pretrained weights matter.}
A random initialization of TimesFM never helps, and it loses 9.7 points on Texas, 3.2 on
Roman-empire, and 2.4 on Cora (\cref{fig:node}a).
Removing TimesFM loses 8.8 points on Roman-empire, 3.8 on Texas, and 1.2 on Amazon-ratings.
The parameter-free path-EMA encoder recovers part of this gap but stays 2.4 and 2.7 points below
TimesFM on Roman-empire and Texas.
On Cora and PubMed, however, the signals alone score 1.0 and 1.5 points higher.
On these homophilous datasets the gain of \ours{} comes from the readout, not from the encoder.
The pretrained encoder pays off where the order along a path carries information, as on
Roman-empire and Texas.
Pretraining on time series gives the encoder a prior over how a value depends on the values before
it.
The parallel flows present the neighbors of a node in this ordered form, so the encoder can use
the prior although it has never seen a graph.

\textbf{The gain grows with the length of the sequences.}
If the gain of the pretrained weights comes from a prior over sequences, it should grow with the
length of the sequences.
We cut every path into windows of at most 8 to 480 nodes, which overlap by a quarter of their
length, up to 64 nodes, so that every node keeps predecessors inside a window.
On Roman-empire the gain rises from 1.4 points with windows of 8 nodes to 2.0, 2.7, and 3.2 points
at 32, 128, and 480 nodes, while on Amazon-ratings, whose edges follow no sequence, it stays between
0.6 and 0.9 points (\cref{fig:node}b).
A longer ordered context thus helps only an encoder that has learned how sequences evolve.
Because the random initialization has the same architecture and context length, the growing gain
comes from what TimesFM learned in pretraining and not from its capacity.

\textbf{The gain holds at every layer.}
Reading the tokens at layers 5, 10, 15, and 20 of the 20-layer TimesFM, the pretrained weights add
1.3 to 2.4 points on Cora and 1.9 to 3.3 points on Roman-empire (\cref{fig:node}c).
The pretrained encoder is stable from layer 5 to layer 15, within 0.2 points on Cora and 0.6 points
on Roman-empire, and it loses 1.7 points on Roman-empire at the last layer, whose states are shaped
most by the forecasting head.
The result therefore does not depend on the choice of layer, and the gain of the pretrained weights
appears already in the lower layers.
One fixed layer thus serves every dataset, and \ours{} needs no layer selection on the target
graph.

\subsection{What Parallel Flows Add}
\label{sec:ana-flows}

\textbf{Ordered context along the paths.}
To isolate the flows, we encode every node as a sequence of length one and keep the readout
unchanged, including its propagation over the graph.
We also shuffle the nodes inside every path and cover only half of the edges, which test whether
the order along a path and the coverage of all edges matter.
Encoding every node alone removes 9.3 points on Roman-empire, 7.6 on Texas, and 1.0 on CiteSeer (\cref{fig:node}a).
Shuffling the nodes inside every path loses 8.2 points on Roman-empire and 5.9 on Texas, nearly as
much as removing the flows, so the order of the nodes along a path matters.
Covering only half of the edges loses 3.6 points on Roman-empire and 3.8 on Texas, so the flows
help because they cover every edge.
On these two heterophilous datasets, the ordered context of the flows therefore carries information
that the neighbor averaging of the readout cannot replace.
On PubMed and Amazon-ratings, however, encoding each node alone scores 0.7 and 2.3 points higher,
so the path context adds noise there.

\begin{figure}[t]
  \centering
  \includegraphics[width=\columnwidth]{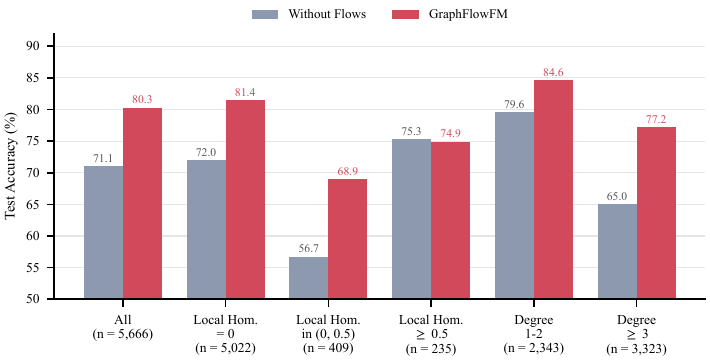}
  \caption{Roman-empire, test accuracy (\%) of \ours{} with and without parallel flows, by local homophily (the share of a node's neighbors in its own class) and by degree.}
  \label{fig:roman}
\end{figure}

\textbf{Where the flows help, a case study on Roman-empire.}
\label{sec:exp-case}
Roman-empire is the dataset on which parallel flows matter most.
\cref{fig:roman} compares \ours{} with the same model in which every node is encoded alone.
The flows raise test accuracy from 71.1\% to 80.3\%, and they correct 675 test nodes while they
break 153.
Of the 5,666 test nodes, 5,022 have no neighbor of their own class, and on these nodes the flows
add 9.4 points.
On the 409 nodes with a minority of same-class neighbors they add 12.2 points, and on the 235 nodes
whose neighbors mostly share their class they change accuracy by $-0.4$ points.
The gain also grows with degree, from 5.0 points for nodes of degree one or two to 12.2 points for
nodes of degree three or more, which lie on more paths.
For a node whose neighbors all belong to other classes, neighbor averaging helps only through the
mix of classes around the node, while an ordered path also tells the encoder which word comes
before the node.
Nodes without same-class neighbors are where averaging over neighbors fails, and they explain why
\ours{} leads the strongest baseline by 10.0 points on Roman-empire (\cref{tab:main}).

\begin{figure}[t]
  \centering
  \includegraphics[width=\columnwidth]{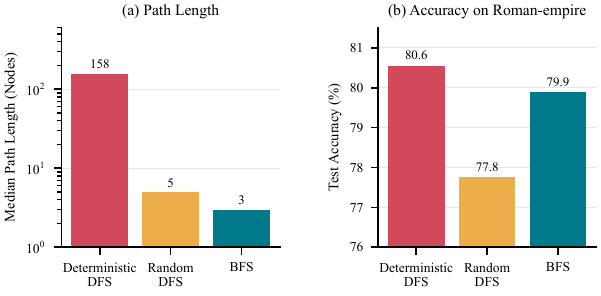}
  \caption{Paths produced by three traversals on Roman-empire, all of which cover every edge with 7 or 8 flows. (a) Median length of the paths. (b) Test accuracy.}
  \label{fig:traversal}
\end{figure}

\begin{figure*}[t]
  \centering
  \includegraphics[width=\textwidth]{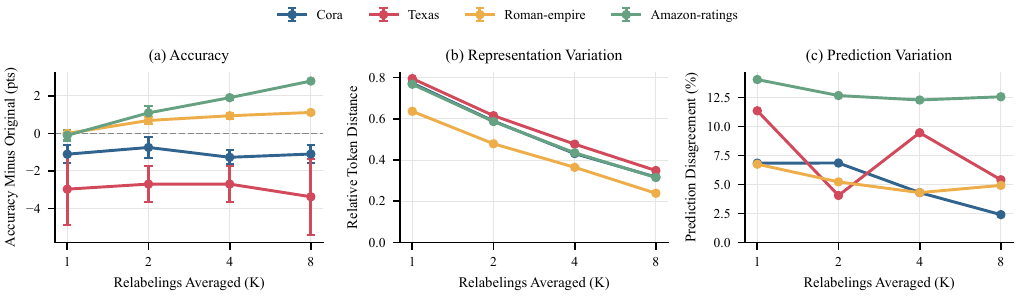}
  \caption{Accuracy, token distance, and prediction disagreement against the number $K$ of random vertex relabelings. Each relabeling randomly permutes the vertex ids and gives one decomposition, and the fused tokens are averaged over $K$ relabelings before the readout, with the label splits fixed. $K = 1$ is a single relabeling. (a) Change in test accuracy relative to the original vertex ids. Error bars show the standard deviation over disjoint sets of $K$ relabelings. (b) Relative distance $\|Z_1 - Z_2\|_F / \sqrt{\|Z_1\|_F \|Z_2\|_F}$ between the averaged tokens $Z_1$ and $Z_2$ of two disjoint sets of $K$ relabelings, computed on 256-dimensional projections. (c) Share of test nodes whose predictions differ between two disjoint sets of $K$ relabelings.}
  \label{fig:relabel}
\end{figure*}

\textbf{A consistent order matters more than long paths.}
On Roman-empire, whose edges link consecutive and syntactically dependent words, the deterministic
DFS of \cref{alg:flows} follows the word sequence, and half of the nodes lie on paths of at least
158 nodes.
\cref{fig:traversal} compares it with a DFS with random roots and neighbor order and with BFS
forests, which tests whether long paths or a consistent order of the nodes drive the accuracy.
All three cover every edge with 7 or 8 flows, but the random DFS and BFS split the same edges into
two to three times as many short paths.
The random DFS cuts across the word sequence and loses 2.8 points, while BFS, which gives the
shortest paths, loses only 0.7 points.
Both deterministic traversals visit the neighbors of a node in a fixed order, so neighboring words
appear in a consistent order along the paths, and the random DFS loses this consistency.
The consistency of the order thus matters more than the length of the paths, and long paths add
about one point on top (\cref{fig:node}b).
On Amazon-ratings neither traversal changes accuracy by more than 0.8 points.
The deterministic DFS of \ours{} fixes the visiting order by degree, so it provides this
consistency without any learned ordering.

\begin{figure}[!htb]
  \centering
  \includegraphics[width=\columnwidth]{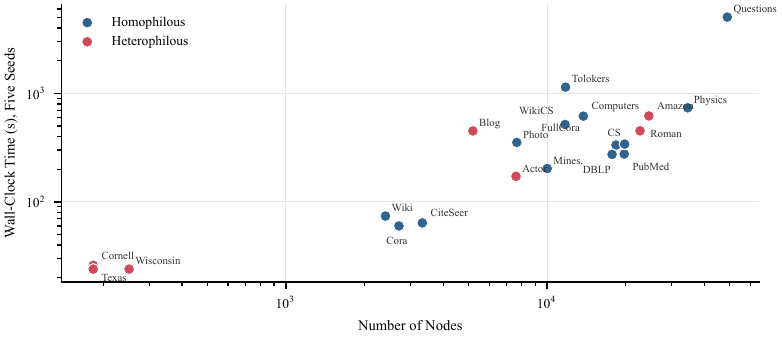}
  \caption{Wall-clock time of one complete run with five seeds on one RTX 4090D, including the decomposition, the encoding of both groups, and the readout, against the number of nodes. All 21 datasets take 3.3 GPU hours in total. For reference, GraphAny reports hyperparameter searches of 4 GPU hours for GraphAny, 240 for MLP, GCN, and GAT together, and 672 for ACM-GNN \citep{zhao2025graphany}.}
  \label{fig:time}
\end{figure}

\subsection{Sensitivity to the Vertex Order}
\label{sec:ana-order}

\label{app:relabel}

The DFS of \cref{alg:flows} visits vertices of equal degree in the order of their ids, so the
decomposition depends on the arbitrary numbering of the vertices.
To measure this dependence, we relabel the vertices of Cora, Texas, Roman-empire, and
Amazon-ratings with ten random permutations and decompose each relabeled graph.
We then map the sequences back to the original vertices and keep the encoder, the readout, and the
label splits unchanged.

\begin{table}[!ht]
\centering
\caption{Sensitivity to the vertex order. Test accuracy (\%) with the original vertex ids and over ten random relabelings of each dataset, each giving one deterministic decomposition, with the label splits fixed and the predictions mapped back to the original nodes. Disagreement is the share of test nodes whose prediction differs between two relabelings, averaged over all pairs. Original differs from \cref{tab:main} on Texas because the two splits are weighted equally here, whereas the five seeds of \cref{tab:main} use the first split three times.}
\label{tab:relabel}
\footnotesize
\setlength{\tabcolsep}{4pt}
\begin{tabular}{lcccc}
\toprule
Dataset & Original & Relabeled & Range & Disagreement \\
\midrule
Cora & 83.1 & 82.0 $\pm$ 0.5 & 81.2--82.6 & 7.0\% \\
Texas & 83.8 & 80.8 $\pm$ 1.9 & 78.4--83.8 & 10.1\% \\
Roman-empire & 80.6 & 80.6 $\pm$ 0.2 & 80.4--80.9 & 7.0\% \\
Amazon-ratings & 50.8 & 50.6 $\pm$ 0.3 & 50.2--51.0 & 14.4\% \\
\bottomrule
\end{tabular}
\end{table}

On Roman-empire and Amazon-ratings, the original ids score at most 0.1 points above the mean over
relabelings, whose standard deviation is at most 0.3 points (\cref{tab:relabel}).
A relabeling changes the order among vertices of equal degree, and with it the paths, but it keeps
the roots and neighbors ordered by degree.
The random DFS of \cref{sec:ana-flows}, which drops this degree order, loses 2.8 points on
Roman-empire, so the degree order appears to matter more than the particular paths.
On Cora and Texas, the original ids score 1.1 and 3.0 points above this mean.
The original vertex order of these two datasets is thus favorable, and the mean over relabelings
is the more conservative estimate of the accuracy of \ours{}.
Even where the mean accuracy is stable, 7.0 to 14.4\% of the test predictions differ between two
relabelings.

Averaging over several decompositions reduces this variation.
We average the fused tokens over $K = 2$, 4, and 8 relabelings and compare two disjoint sets of $K$
relabelings (\cref{fig:relabel}).
The relative distance between their tokens falls from 0.64--0.80 at $K = 1$ to 0.24--0.35 at
$K = 8$, close to the $1/\sqrt{K}$ rate of independent averages.
At $K = 8$, averaging raises accuracy over the original ids by 1.1 points on Roman-empire and 2.8
points on Amazon-ratings.
On Cora and Texas it does not raise accuracy, which stays near the mean over single relabelings.
Since averaging multiplies the encoding time by $K$, \ours{} encodes each graph once and leaves
averaging as an option.
For graph-level tasks, \cref{prop:single} shows that one decomposition per graph, drawn from a
random vertex order, gives an unbiased and consistent estimate of the risk averaged over vertex
orders.

\subsection{Feature and Structure Groups}
\label{sec:ana-groups}

Removing one group at a time tests what each group contributes.
The structure group alone is far below the full model on most datasets with informative features,
for example 37.5 against 80.6 on Roman-empire, and within three points only on Minesweeper and
Amazon-ratings.
Removing the structure group costs 2.7 points on Texas and 0.9 on Roman-empire and changes the
homophilous datasets by at most 0.3 points.
The alignment weights explain this pattern (\cref{fig:node}d and \cref{app:align}).
They give the structure tokens about a third of the weight on the WebKB datasets Cornell and Texas,
where the structure group helps most, and at most 10\% on Minesweeper and Cora, where it does not
help.
Because the readout sets these weights from the training labels of each dataset, \ours{} uses the
structure group where it helps and keeps it from degrading accuracy where it would only add noise.
On link prediction and graph classification the structure group carries more, as on GitHub and
AirUS, where \ours{} uses it alone (\cref{sec:exp-link}), and on HIV, where it reaches 73.1 ROC-AUC
without any atom type (\cref{sec:exp-graph}).

\subsection{Hyperparameters and Computational Time}
\label{sec:exp-impl}

Every hyperparameter of \ours{} is set once and applied to all datasets and tasks
(\cref{tab:hparams} in \cref{app:impl}).
What adapts to a dataset is the closed-form fit of each candidate on the training labels and the
greedy ensemble of candidates on the validation labels, so the kernels, propagations, and ridge
strengths that enter the final prediction differ between datasets.
Sequences are sorted by length and packed into batches of at most 150,000 tokens, and hidden states
are accumulated per node, so memory grows with the number of nodes and not with the number of
occurrences.
The readout solves the full kernel ridge problem without approximation on every graph, including
Questions with 48,921 nodes.
\cref{fig:time} shows the wall-clock time of one complete run per graph.
All 21 datasets take 3.3 GPU hours in total, of which Questions takes 1.4.
For reference, GraphAny reports hyperparameter searches of 4 GPU hours for GraphAny, 240 for MLP,
GCN, and GAT together, and 672 for ACM-GNN \citep{zhao2025graphany}.
\ours{} fixes its hyperparameters once and needs no such search on the target graph.

\FloatBarrier
\raggedbottom
\section{Conclusion}
\label{sec:conclusion}

We have shown that a frozen time-series foundation model that has never been trained on a graph can
serve as a graph foundation model once the graph is decomposed into parallel flows.
On 21 datasets of node classification, \ours{} attains the best average rank among graph foundation models
published in 2025 and 2026, with its largest margins on heterophilous datasets.
With the same frozen encoder and no gradient update, \ours{} also attains the highest ROC-AUC on
five of six link prediction datasets, ahead of pre-trained and fine-tuned graph foundation models,
and ranks second in molecule classification, behind only a model pre-trained on molecules.

\bibliographystyle{IEEEtranN}
\bibliography{references}

\clearpage
\onecolumn
\appendices
\crefalias{section}{appendix}
\crefalias{subsection}{appendix}
\setcounter{figure}{0}
\setcounter{table}{0}
\setcounter{equation}{0}
\renewcommand{\thefigure}{S\arabic{figure}}
\renewcommand{\thetable}{S\arabic{table}}
\renewcommand{\theequation}{S\arabic{equation}}
\renewcommand{\theHfigure}{S\arabic{figure}}
\renewcommand{\theHtable}{S\arabic{table}}
\renewcommand{\theHequation}{S\arabic{equation}}
\makeatletter
\setlength{\@fptop}{0pt}
\renewcommand{\topfraction}{0.9}
\renewcommand{\bottomfraction}{0.8}
\renewcommand{\textfraction}{0.07}
\renewcommand{\floatpagefraction}{0.8}
\makeatother

\section{Datasets and Protocols}
\label{app:protocols}

\subsection{Node Classification}
\label{app:data}

\begin{table*}[!ht]
\centering
\caption{The 21 node classification datasets, taken with their splits from GraphAny \citep{zhao2025graphany}. Edges are undirected, and Hom.\ is the edge homophily.}
\label{tab:data}
\small
\setlength{\tabcolsep}{4pt}
\begin{tabular}{llllrrrrr}
\toprule
Group & Dataset & Source & Domain & Nodes & Edges & Feat. & Classes & Hom. \\
\midrule
Homophilous & Cora & Planetoid & citation & 2,708 & 5,278 & 1,433 & 7 & 0.81 \\
 & CiteSeer & Planetoid & citation & 3,327 & 4,552 & 3,703 & 6 & 0.74 \\
 & PubMed & Planetoid & citation & 19,717 & 44,324 & 500 & 3 & 0.80 \\
 & Photo & Amazon (Shchur et al.) & co-purchase & 7,650 & 119,081 & 745 & 8 & 0.83 \\
 & CS & Coauthor (Shchur et al.) & co-authorship & 18,333 & 81,894 & 6,805 & 15 & 0.81 \\
 & Computers & Amazon (Shchur et al.) & co-purchase & 13,752 & 245,861 & 767 & 10 & 0.78 \\
 & Physics & Coauthor (Shchur et al.) & co-authorship & 34,493 & 247,962 & 8,415 & 5 & 0.93 \\
 & DBLP & CitationFull & citation & 17,716 & 52,867 & 1,639 & 4 & 0.83 \\
 & FullCora & CitationFull & citation & 19,793 & 63,421 & 8,710 & 70 & 0.57 \\
 & Wiki & AttributedGraph & web & 2,405 & 11,596 & 4,973 & 17 & 0.61 \\
 & WikiCS & WikiCS & web & 11,701 & 215,603 & 300 & 10 & 0.65 \\
 & Minesweeper & Platonov et al. & synthetic grid & 10,000 & 39,402 & 7 & 2 & 0.68 \\
 & Tolokers & Platonov et al. & crowdsourcing & 11,758 & 519,000 & 10 & 2 & 0.59 \\
 & Questions & Platonov et al. & social & 48,921 & 153,540 & 301 & 2 & 0.84 \\
\midrule
Heterophilous & Roman-empire & Platonov et al. & text & 22,662 & 32,927 & 300 & 18 & 0.05 \\
 & Amazon-ratings & Platonov et al. & co-purchase & 24,492 & 93,050 & 300 & 5 & 0.38 \\
 & BlogCatalog & AttributedGraph & social & 5,196 & 171,743 & 8,189 & 6 & 0.40 \\
 & Cornell & WebKB & web & 183 & 277 & 1,703 & 5 & 0.12 \\
 & Texas & WebKB & web & 183 & 279 & 1,703 & 5 & 0.06 \\
 & Wisconsin & WebKB & web & 251 & 450 & 1,703 & 5 & 0.18 \\
 & Actor & Geom-GCN & co-occurrence & 7,600 & 26,659 & 932 & 5 & 0.22 \\
\bottomrule
\end{tabular}
\end{table*}

\cref{tab:data} lists the 21 datasets from GraphAny that we evaluate.
Edges are counted after symmetrization and removal of self-loops, and edge homophily is the share
of edges whose two endpoints share a class.
The feature dimension ranges from 7 to 8,710, the number of classes from 2 to 70, and the size from
183 to 48,921 nodes.
\ours{} handles all of them with the same frozen encoder, and it turns the features of every
dataset into 32 feature variates by the same fixed weighting and truncated SVD (\cref{app:impl}).

\textbf{Splits and seeds.}
We follow the split rule of the GraphAny code, so the seed affects only the split.
The datasets fall into three groups by the splits they come with.
Cora, CiteSeer, and PubMed come with a single public split, which every seed uses.
Ten datasets, Cornell, Texas, Wisconsin, Actor, Roman-empire, Amazon-ratings, Minesweeper,
Tolokers, Questions, and WikiCS, come with several official splits.
Seed $s$ uses split $s \bmod 2$, so the five seeds alternate between the first two of them.
For the other eight datasets, every seed draws a new stratified split.
\ours{} itself involves no randomness, because the encoder is frozen, the DFS order is fixed, and
the readout and the ensemble selection are deterministic.
Its standard deviation in \cref{tab:full} therefore reflects only the variation across splits.
It is zero on the three single-public-split datasets, and on Cornell, where both splits give 27 correct
predictions out of 37 test nodes.

\subsection{Link Prediction}
\label{app:link-protocol}

For each graph and seed, the undirected edges are split at random into 70\% training, 15\%
validation, and 15\% test edges.
As negative examples, each split also gets the same number of randomly sampled node pairs that are
not connected, which we call non-edges.
The training, validation, and test non-edges do not overlap.
The parallel flows, both signal groups, and the propagation are computed from the training edges
only, so no validation or test edge enters any representation.
The readout is fitted on at most 10,000 training edges and as many training non-edges, and the
validation and test scores use at most 20,000 edges and 20,000 non-edges each, all drawn at random
per seed.
The node features of AirUS are one-hot node identifiers, and GitHub has no informative node
features, so \ours{} uses only the structure group on these two graphs.
We report ROC-AUC averaged over five seeds.

\subsection{Graph Classification}
\label{app:graph}

We use the scaffold splits of five OGB molecule benchmarks, HIV, BBBP, BACE, ToxCast, and Tox21,
as in GIT \citep{wang2025git}.
The molecules of a benchmark are decomposed as one disjoint union of graphs, so no path crosses
two molecules.
The feature group holds 32 SVD variates of the one-hot atom features, and the structure group holds
the eight structural variates of each atom, computed within its molecule.
A molecule is represented by the mean of the propagated representations of its atoms.
For each task, the readout is fitted on the training molecules labeled for that task, and the
validation and test scores use only labeled molecules.
A task whose validation or test molecules all belong to one class has no ROC-AUC and is skipped,
and a benchmark with several tasks is scored by the mean ROC-AUC over the remaining tasks.
Since the scaffold split is fixed and \ours{} is deterministic, each benchmark gives a single
result.

\section{Implementation Details}
\label{app:impl}

\cref{tab:hparams} lists every hyperparameter of \ours{}.
Each is set once for all datasets and tasks, and none is selected on the target graph.

\begin{table}[!htbp]
\centering
\caption{Hyperparameters of \ours{}, shared by all datasets.}
\label{tab:hparams}
\small
\begin{tabular}{lll}
\toprule
Component & Hyperparameter & Value \\
\midrule
Node signals & feature group & 32 SVD variates of the weighted features (\cref{sec:encode}) \\
 & structure group & $\log(1+d_v)$, random-walk return probabilities for 1 to 7 steps \\
Parallel flows & DFS order & roots by decreasing degree, neighbors by increasing degree \\
 & window, overlap & 480 nodes, 64 nodes \\
 & directions & both \\
Encoder & model & TimesFM-3, frozen, 20 layers, width 1,280 \\
 & patch length $P$ & 32 \\
 & layer & 15 \\
 & input normalization & identity \\
Readout & blocks & feature and structure self and context tokens, feature signals \\
 & representations & five fixed block sets and one alignment-weighted set \\
 & kernels & linear, Gaussian with median bandwidth \\
 & propagations & $\hat A^0$, $\hat A^1$, $\hat A^2$, PPR with $\alpha=0.1$, $T=10$ \\
 & ridge $\lambda$ & 0.1, 1, 10 (times the mean kernel diagonal), class-prior intercept \\
 & softmax temperature & $3/\sigma_F$ \\
 & prediction smoothing & 10 steps, $\alpha=0.1$, node targets only \\
 & ensemble selection & 30 greedy steps with replacement, validation accuracy or ROC-AUC \\
\bottomrule
\end{tabular}
\end{table}

\textbf{Feature weighting.}
The weighting depends on the type of the features.
Binary features are weighted as $\log(1+x_{vj})\,\mathrm{idf}_j$ with
$\mathrm{idf}_j=\log\frac{1+n}{1+n_j}+1$, where $n_j$ is the number of nodes with feature $j$.
Other non-negative features, such as TF-IDF weights, are kept as given.
Both kinds are then normalized to unit $\ell_2$ norm per node.
Features with negative entries, such as text embeddings, are divided per node by the sum of their
absolute values.
The signal matrix $S_{\mathrm{f}}$ holds the projections of the weighted features onto their top
$p_{\mathrm{f}}=32$ right singular vectors.

\textbf{Degenerate inputs.}
A dataset with $d \le 32$ features keeps $d-1$ SVD components, and the remaining variates are set
to zero, which affects Minesweeper ($d=7$) and Tolokers ($d=10$).
The SVD is a randomized truncated SVD with a fixed seed, so the signals are deterministic.
A node without features keeps a zero row, since the normalization divides only by a positive norm.
A column with zero variance becomes zero after standardization.
On a graph without self-loops, the one-step return probability is zero for every node, so this
variate becomes zero and seven structural variates remain informative.
An isolated node has degree zero and zero return probabilities.

\textbf{Number of structural variates.}
We use eight structural variates, although the encoder accepts 32.
Return probabilities of adjacent walk lengths have a correlation of about 0.95.
If we compute return probabilities for 32 walk lengths, their first three to five principal
components already explain 99\% of their variance, depending on the dataset.
The later walk lengths therefore add little information, and eight variates keep the token block
small.

\textbf{Batching and memory.}
Sequences are sorted by length, packed into batches of at most 150,000 tokens, and left-padded with
masks.
A forward hook captures the hidden states at layer 15 and accumulates them per node, so memory grows
with the number of nodes and not with the number of occurrences.
On large graphs, the nodes are split into several groups, and the encoder runs once per group and
keeps only the tokens of that group.
For training sets above 15,000 nodes, the ridge system is solved by a Cholesky factorization that
overwrites the kernel matrix.

\section{Full Results}
\label{app:results}

\subsection{Node Classification}
\label{app:full}

\begin{table*}[!ht]
\centering
\caption{Node classification accuracy (\%) on the 21 datasets under the GraphAny protocol. \ours{} is reported as mean $\pm$ standard deviation over five seeds. \ours{} is deterministic, so its standard deviation reflects only the variation across splits and is zero where all seeds share one split or the splits give the same accuracy (\cref{app:data}). GraphAny-Arxiv is GraphAny trained on ogbn-arxiv, RGVT-Linear is RGVT with a linear head, and TAG is not yet formally published and is listed for reference. n/r means not reported. Best in bold, second underlined.}
\label{tab:full}
\scriptsize
\setlength{\tabcolsep}{3pt}
\begin{tabular}{lccccccccc}
\toprule
Dataset & GraphAny & GraphAny-Arxiv & TS-Mean & TS-GAT & NodePFN & RGVT & RGVT-Linear & TAG & \ours{} \\
\midrule
\multicolumn{10}{l}{\emph{Homophilous}} \\
Cora & 77.82 & 79.38 & n/r & n/r & 82.06 & 81.68 & 81.32 & \textbf{83.50} & \underline{83.10$\pm$0.00} \\
CiteSeer & 67.50 & 68.34 & 68.66 & 68.70 & 67.30 & 70.12 & \textbf{71.96} & 68.08 & \underline{70.40$\pm$0.00} \\
PubMed & 77.46 & 76.36 & 74.98 & 75.46 & 78.00 & 78.20 & \underline{79.02} & 78.96 & \textbf{79.30$\pm$0.00} \\
Photo & 90.18 & 90.60 & 90.18 & 89.66 & 90.53 & \underline{92.29} & 92.02 & 89.63 & \textbf{92.47$\pm$0.66} \\
CS & 90.85 & 90.45 & 90.92 & 91.13 & 91.55 & \underline{92.28} & 91.32 & 90.49 & \textbf{93.04$\pm$0.29} \\
Computers & 82.00 & 83.04 & 81.37 & 80.23 & 81.42 & 84.14 & n/r & \underline{84.33} & \textbf{86.63$\pm$0.63} \\
Physics & 92.54 & 92.69 & 92.61 & 92.23 & \textbf{93.43} & \underline{93.18} & n/r & 92.31 & 92.93$\pm$0.38 \\
DBLP & 70.13 & 70.90 & 66.42 & 64.30 & 74.71 & \textbf{78.12} & n/r & 71.92 & \underline{77.20$\pm$0.94} \\
FullCora & 56.29 & 57.25 & 53.58 & 52.82 & n/r & \textbf{63.26} & n/r & 54.56 & \underline{60.84$\pm$0.40} \\
Wiki & 61.10 & 62.96 & 69.89 & \underline{72.53} & n/r & 72.26 & n/r & 70.09 & \textbf{75.16$\pm$1.14} \\
WikiCS & 73.77 & 74.95 & 74.16 & 73.24 & 75.98 & 78.92 & n/r & \underline{79.07} & \textbf{79.91$\pm$0.47} \\
Minesweeper & 80.13 & 80.30 & 80.68 & 80.72 & 80.66 & 80.00 & n/r & \textbf{85.83} & \underline{82.92$\pm$0.10} \\
Tolokers & 78.24 & 78.18 & 78.12 & 78.01 & 78.61 & 78.32 & n/r & \textbf{82.82} & \underline{80.52$\pm$0.50} \\
Questions & 97.11 & 97.09 & 97.02 & 97.02 & 97.02 & 97.13 & n/r & \underline{97.14} & \textbf{97.26$\pm$0.04} \\
\midrule
\multicolumn{10}{l}{\emph{Heterophilous}} \\
Roman-empire & 64.06 & 64.25 & 66.36 & 67.76 & n/r & 70.53 & 67.71 & \underline{74.12} & \textbf{80.55$\pm$0.35} \\
Amazon-ratings & 42.57 & 42.74 & 42.27 & 41.92 & 44.68 & \underline{45.58} & 42.98 & 44.34 & \textbf{50.73$\pm$0.19} \\
BlogCatalog & 77.69 & 73.63 & 76.30 & 78.44 & n/r & \underline{81.39} & n/r & 79.77 & \textbf{90.32$\pm$0.45} \\
Cornell & 66.49 & 65.94 & 68.65 & 71.35 & 71.89 & \underline{74.05} & n/r & \textbf{75.14} & 72.97$\pm$0.00 \\
Texas & 73.51 & 72.97 & 73.51 & 74.05 & 76.22 & 78.92 & n/r & \underline{81.08} & \textbf{84.32$\pm$2.65} \\
Wisconsin & 71.77 & 65.10 & 61.18 & 65.88 & 79.22 & 78.04 & n/r & \textbf{83.14} & \underline{81.18$\pm$5.76} \\
Actor & 29.51 & 28.60 & 28.09 & 28.80 & \underline{32.99} & 32.68 & n/r & 31.18 & \textbf{33.43$\pm$0.06} \\
\midrule
Avg.\ rank (21) & 6.10 & 6.00 & 6.35 & 6.10 & 4.06 & 2.95 & 3.57 & 3.48 & 1.52 \\
Ranked first & 0/21 & 0/21 & 0/20 & 0/20 & 1/17 & 2/21 & 1/7 & 5/21 & 12/21 \\
\bottomrule
\end{tabular}
\end{table*}

\cref{tab:full} reports every dataset with the standard deviation of \ours{} and every reported
variant of the baselines.
Compared with \cref{tab:main}, it adds GraphAny trained on ogbn-arxiv, RGVT with a linear head,
which reports only seven of the datasets, and TAG \citep{hayler2025tag}, which is listed for
reference because it is not yet formally published.
Over all 21 datasets and nine methods and variants, \ours{} keeps the best average rank, 1.52,
against 2.95 for RGVT and 3.48 for TAG, and it ranks first on 12 datasets.
TAG, which feeds aggregated features to a tabular foundation model, ranks first on five datasets,
Cora, Minesweeper, Tolokers, Cornell, and Wisconsin.

\subsection{Link Prediction}
\label{app:link}

\begin{table*}[!ht]
\centering
\caption{Link prediction (ROC-AUC, \%). \emph{Graph Training} states whether a method is trained on graph data: on each target graph (Per-Dataset), or by pre-training on a graph corpus followed by fine-tuning on the target graph (Pre-Train + FT) or without fine-tuning (Pre-Train, No FT). Baseline numbers are copied from \citet{liu2026rgfm} (first block), \citet{sun2026gauge} (second block), \citet{wang2025git} (third block) and \citet{sun2025riemanngfm} (fourth block); ``--'' means not reported. A method reported more than once (GCOPE, SAMGPT, MDGFM, GCN, and OFA) appears once, in the block of its first source, with the highest reported value on each dataset. GraphFlowFM makes no gradient update.}
\label{tab:link}
\small
\begin{tabular}{llcccccc}
\toprule
Method & Graph Training & Cora & CiteSeer & PubMed & AirUS & Ratings & GitHub \\
\midrule
GCOPE & Pre-Train + FT & 88.71 & 88.60 & 90.84 & 86.17 & -- & 82.16 \\
SAMGPT & Pre-Train + FT & 88.62 & 89.91 & 84.92 & -- & -- & -- \\
MDGFM & Pre-Train + FT & 88.41 & 89.12 & 84.78 & -- & -- & -- \\
R-GFM & Pre-Train + FT & 89.27 & 90.88 & 88.62 & -- & -- & -- \\
\midrule
GFT & Pre-Train, No FT & -- & -- & 55.83 & -- & -- & -- \\
RAGraph & Pre-Train, No FT & -- & -- & 58.29 & -- & -- & -- \\
GET & Pre-Train, No FT & -- & -- & 48.82 & -- & -- & -- \\
G2PM & Pre-Train, No FT & -- & -- & 53.08 & -- & -- & -- \\
UniPrompt & Pre-Train, No FT & -- & -- & 56.32 & -- & -- & -- \\
GAUGE & Pre-Train, No FT & -- & -- & 64.03 & -- & -- & -- \\
\midrule
GCN & Per-Dataset & 87.34 & 90.70 & 91.16 & 92.37 & 97.77 & 87.48 \\
BGRL & Per-Dataset & 83.96 & 81.51 & 84.01 & -- & 98.05 & -- \\
GraphMAE & Per-Dataset & 85.57 & 84.55 & 89.83 & -- & 95.35 & -- \\
OFA & Pre-Train + FT & 82.82 & 82.62 & 92.26 & -- & 97.79 & -- \\
GIT-G & Pre-Train + FT & 87.79 & 87.59 & 84.35 & -- & 98.45 & -- \\
GIT-S & Pre-Train + FT & 88.58 & 88.50 & 87.78 & -- & 98.49 & -- \\
\midrule
GraphSAGE & Per-Dataset & -- & 87.29 & 87.02 & 92.17 & -- & 79.13 \\
DGI & Per-Dataset & -- & 96.90 & 88.39 & 92.50 & -- & 86.39 \\
GraphMAE2 & Per-Dataset & -- & 92.75 & 89.46 & 88.23 & -- & 87.11 \\
OpenGraph & Pre-Train + FT & -- & 76.78 & 70.02 & 85.32 & -- & 86.72 \\
LLaGA & Pre-Train + FT & -- & 86.26 & 84.04 & 77.90 & -- & 71.25 \\
\midrule
\textbf{GraphFlowFM} & \textbf{No Gradient Update} & \textbf{92.86} & \textbf{93.81} & \textbf{95.89} & \textbf{95.78} & \textbf{99.70} & \textbf{91.69} \\
\bottomrule
\end{tabular}
\end{table*}

\cref{tab:link} lists all baseline results that we compare with, taken from the published papers and
grouped into four blocks by source.
These sources use their own edge splits, so the comparison is between published results on the same
datasets.

\section{Further Analyses}
\label{app:analyses}

\subsection{Ablations}
\label{app:ablation}

\begin{figure}[!htbp]
  \centering
  \includegraphics[width=0.72\textwidth]{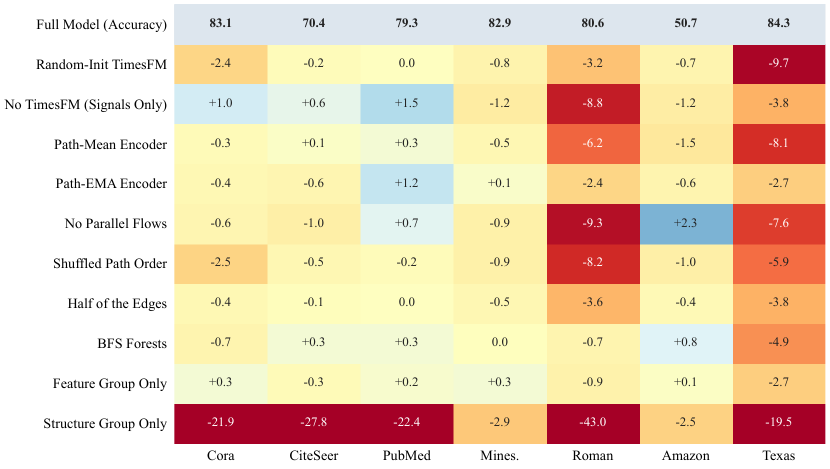}
  \caption{Ablations on seven representative datasets. The first row is the test accuracy (\%) of the full model, averaged over five seeds, and every other row is the change in accuracy from it. }
  \label{fig:ablation-full}
\end{figure}

\cref{fig:node}a shows only how much the accuracy changes when one component of
\ours{} is changed.
\cref{fig:ablation-full} gives the full numbers on the seven datasets of that figure.
Its first row is the test accuracy of the full model, and each other row is the change in percentage points
when one component is changed.
The changes are of three kinds.
The encoder is replaced by a randomly initialized TimesFM, by the input signals without any
encoder, or by a mean or an exponential moving average along the path (\cref{sec:ana-encoder}).
The parallel flows are removed, shuffled within each path, reduced to half of the edges, or built
from BFS forests (\cref{sec:ana-flows}).
Finally, only the feature group or only the structure group is kept (\cref{sec:ana-groups}).

\subsection{Encoder Depth and Sequence Length}
\label{app:layers}

\begin{figure}[!htbp]
  \centering
  \includegraphics[width=0.9\textwidth]{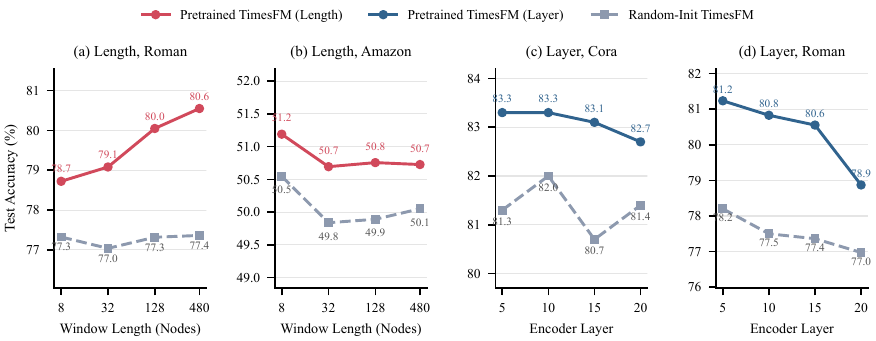}
  \caption{Test accuracy of the pretrained TimesFM and of a randomly initialized TimesFM of the same architecture. (a, b) Every path is cut into windows of at most the given number of nodes, and 480 is the default of \ours{}. (c, d) The tokens are read at different encoder layers, and layer 15 is the default of \ours{}.}
  \label{fig:length-layer}
\end{figure}

\cref{fig:node}b and \cref{fig:node}c show the accuracy of the pretrained TimesFM
minus that of a randomly initialized TimesFM (\cref{sec:ana-encoder}).
\cref{fig:length-layer} gives the accuracies of the two models from which these differences are
computed.
These runs vary only the window length or the layer and keep the rest of \ours{} unchanged.

\subsection{Alignment Weights}
\label{app:align}

\begin{figure}[!htbp]
  \centering
  \includegraphics[width=0.75\textwidth]{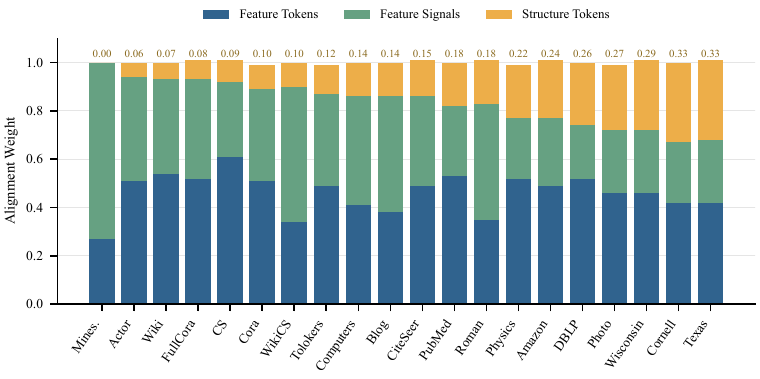}
  \caption{Alignment weights of the feature tokens, the feature signals, and the structure tokens in the alignment-weighted representation. The number above each bar is the share of the structure tokens. Questions uses the low-memory readout, which does not log the weights.}
  \label{fig:align}
\end{figure}

\cref{fig:align} shows how the alignment-weighted representation of \cref{sec:readout} divides its
weight among the feature tokens, the feature signals, and the structure tokens.
The weights are computed from the training labels only.
The structure tokens receive the largest share on the small WebKB datasets, Cornell, Texas, and
Wisconsin, whose labels depend on the role of a web page in the link structure.
They receive no weight on Minesweeper, whose features already count the mines around a cell.

\section{Case Studies}
\label{app:cases}

We examine link prediction on a citation graph and graph classification on molecules.
The node classification case study on Roman-empire is in \cref{sec:ana-flows}.

\textbf{Link prediction on Cora.}
\cref{fig:link} scores the same test pairs with \ours{} and with the Adamic-Adar heuristic.
Like \ours{}, the heuristic sees only the training edges, and it scores each test pair by the
common neighbors of its two nodes in this training graph.
\ours{} reaches 92.7 ROC-AUC, against 67.6 for both Adamic-Adar and common neighbors.
The gap is largest on the 1,298 test pairs whose endpoints share no neighbor in the training
graph.
The heuristics give all of these pairs the same score, whereas \ours{} still reaches 89.2.
Its tokens encode the features along the paths of each node, so two papers can have similar tokens
without a common neighbor.
Pairs that involve hub papers, the top 10\% by degree, are easier for both methods, and \ours{}
reaches 98.7 when both endpoints are hubs.
Of the readout candidates selected by the ensemble, 92\% use the Gaussian kernel and 60\% use one
step of propagation.
As representation, 48\% use the feature self tokens alone and 40\% the alignment-weighted
representation, which weights the feature tokens, the feature signals, and the structure tokens by
their alignment with the labels, as \cref{fig:align} shows for node classification.

\textbf{Graph classification on HIV.}
HIV asks whether a molecule inhibits HIV replication, and only 3.2\% of the 4,113 test molecules
are active.
The readout fits one candidate for each combination of representation, kernel, propagation, and
ridge strength (\cref{sec:readout}).
For each of the six representations, \cref{fig:graph}a reports the test ROC-AUC of the candidate
with the highest validation ROC-AUC, and it also reports the ensemble, which is the final
prediction of \ours{}.
The feature tokens, which encode the atom types along the paths of a molecule, reach 77.3 to 77.6
ROC-AUC alone.
The structure tokens reach 73.1 from the shape of the molecule alone, without any atom type.
The ensemble reaches 78.0, and 92\% of its selected candidates use the feature self tokens, all of
them with the Gaussian kernel.
By size, \ours{} is strongest on molecules with 27 or more atoms (86.9) and weaker on molecules
with 20 to 26 atoms (56.3).

\end{document}